%% file: main.tex
\documentclass[11pt]{article}
\usepackage[a4paper,margin=2.5cm]{geometry}
\usepackage[T1]{fontenc}
\usepackage{lmodern}
\input{glyphtounicode}\pdfgentounicode=1

\usepackage{amsmath,amssymb}
\usepackage{graphicx}
\usepackage{booktabs}
\usepackage{array}
\usepackage{xcolor}
\usepackage{changepage}
\usepackage[section]{placeins}
\usepackage[round,authoryear]{natbib}
\usepackage{pdflscape}
\usepackage{longtable}
\usepackage{float}
\usepackage{xurl}
\usepackage[hidelinks]{hyperref}
\graphicspath{{figs/}}

\newcommand{\AUC}{\mathrm{AUC}}
\newcommand{\pkg}{\texttt{scanpathdecomp}}
\newcommand{\Cs}{C}

\newcommand{\extralength}{0cm}

\title{Characterizing Scanpath Decoding Pipelines with Perturbation Profiles}
\author{Saleh Mozaffari$^{1}$ \and Abolfazl Nadi$^{2,3,*}$}
\date{\small $^{1}$Department of Bioinformatics, Kish International Campus, University of Tehran (Tehran office), Tehran 1417955391, Iran; saleh.mozafari@ut.ac.ir\\
$^{2}$Department of Computer Science, School of Mathematics, Statistics and Computer Science, College of Science, University of Tehran, Tehran 1417614411, Iran; a.nadi@ut.ac.ir\\
$^{3}$BRAINS, Gisma University of Applied Sciences, Potsdam, Germany\\
$^{*}$Corresponding author.}

\begin{document}
\maketitle

\begin{abstract}
\noindent When a scanpath descriptor supports a classification, it is rarely clear what the classification rests on. We propose perturbation profiles. Five transformations each target one stated property of every scanpath (fixation order, the coupling of durations to positions, the duration distribution, orientation and handedness, and fine spatial structure) while preserving a listed set of invariants. The classifier is refit, and the change in participant-held-out AUC is the pipeline's sensitivity to that property. Intervals come from a participant bootstrap that refits the classifier; in simulations they were conservative (99.7\% coverage; standard errors about 1.8 times the empirical spread). We profiled nine descriptors on thirteen binary problems from three public datasets, with a within-fold AUC and class-balanced folds and a stated rule for which sensitivities to interpret. Of 565 intervals, 95 excluded zero; 80 were robust to the pooled AUC, 20 fold partitions and longer runs (36 of them within $\pm 0.05$ AUC), and 45 also held with boosted trees. Most came from GazeBase (322 participants). Permuting fixation order raised the AUC of a descriptor based on mean saccade amplitude, because permuted amplitude has as its expectation the mean pairwise distance, a measure of spread. A ``comprehension'' contrast pooled over two programs was a contrast between their layouts: it vanished among trials of the same program, and synthetic gaze drawn from the layout alone reproduced the program contrast beneath it. For native versus non-native readers, spatial jitter at the scale of text lines reduced recurrence's discrimination. A Python package is released.
\end{abstract}

\noindent\textbf{Keywords:} eye tracking; scanpath analysis; perturbation analysis; surrogate data; recurrence quantification analysis; cross-validation; reading; program comprehension

\section{Introduction}
\input{sec_intro_a}

The transformations preserve more than their names suggest: permuting durations within a trial preserves every marginal duration statistic, and a rigid rotation preserves every fixation's eccentricity. We state each transformation's invariants exactly, demonstrate the two blind spots with planted counterexamples, and add a second duration transformation that destroys the marginal. The three datasets were chosen to differ in task, stimulus and label: SB-SAT \citep{ahn2020towards} (95 participants reading SAT passages, with native language and comprehension), EMIP \citep{bednarik2020emip} (programmers reading two short programs, with self-rated expertise and answer correctness; we use its corrected fixation release, \citealp{almadi2021emiptoolkit}) and GazeBase \citep{griffith2021gazebase} (322 participants on reading, video and game tasks). Their labels vary within participants, between participants or both, and their difficulty runs from ceiling to none. Among the descriptors is a graph representation of a trial on the regions of its stimulus, the Fixation Adjacency Graph, which we generalize from stimulus-specific segmentations to any set of areas of interest and to a stimulus-independent lattice. All data are public, and every analysis can be re-run from the released code.

\input{sec_method}

\subsection{Software}
The method is implemented in \pkg\ (Python, MIT licence). A trial is a small dataclass; a descriptor is any function from a trial to a dictionary of features; \texttt{decompose()} runs the observed evaluation, the transformations with a user-chosen jitter grid, and the refit bootstrap, and returns a table of AUCs, a table of sensitivities (with exact invariances and variance fallbacks flagged), and the per-draw bootstrap AUCs from which every interval can be re-derived. The evaluation is leave-one-participant-out by default; an option selects $k$-fold cross-validation over participants, which we use here (ten folds) because the largest problems have 322 participants and the refit bootstrap repeats the evaluation several thousand times per profile. Options select the pooled or within-fold AUC, class-balanced folds, the seed of the fold partition, and the duration donor pool; the package warns when a pooled AUC is requested for a label that is constant within participants. Built-in descriptors cover recurrence quantification, directed transition graphs on a grid, classic statistics, and the undirected Fixation Adjacency Graph on a stimulus mask, on any set of anchor points, or on a lattice. With $k$ folds, participants are put in a seeded random order of their sorted identifiers and dealt to the folds in turn, so the assignment, and with it every reported number, is the same on any machine. The results reported here come from version 0.6.3 of the package, with scikit-learn 1.7.2. Further versions and the learner settings are in the methods appendix of the Appendix. Analysis code and manuscript text were drafted and revised with the assistance of a large language model (Claude, Anthropic) under the authors' direction; the authors reviewed and tested the code and checked the reported statistics against the stored outputs.

\input{sec_sim}

\section{Datasets, Problems and Descriptors}\label{sec:data}
All three datasets are public and de-identified, and none was collected for this study. Table~\ref{tab:problems} lists the problems. Sample sizes were fixed by the datasets rather than chosen for this study; Section~\ref{sec:sim} reports, at these sizes, the interval widths and the sensitivities that were reliably detected.

\paragraph{SB-SAT.} The Stony Brook SAT reading dataset \citep{ahn2020towards} contains EyeLink 1000 recordings (1000 Hz) of 95 participants reading four SAT passages of five to six pages each, followed by comprehension questions, with fixations exported by the manufacturer's software in screen pixels. A trial is the first visit to a reading page; revisits after paging back are excluded, as are fixations outside the $1024\times768$ px display (0.6\%). Trials need at least five fixations. Two problems were formed: native versus non-native English readers (a between-participant label; 66 and 29 participants), and pages of passages answered with accuracy $\geq 0.8$ versus $\leq 0.4$ (a mixed label; intermediate passages are excluded). The text's line spacing, measured as the spacing of word-level fixation medians on the stimulus pages, is 63 px.

\paragraph{EMIP.} The Eye Movements in Programming dataset \citep{bednarik2020emip} contains SMI recordings of 216 programmers, each reading two short programs (Rectangle and Vehicle) and answering a comprehension question after each. We use its corrected fixation release \citep{almadi2021emiptoolkit}, which contains fixations already assigned to code tokens and corrected for vertical drift for 159 participants, and the dataset's metadata for expertise and answers. The release keeps only fixations that fall on a token, which removes off-code looking before any descriptor is computed. A trial is one program reading. Six problems were formed: Rectangle versus Vehicle for the 112 participants with both programs in the same stimulus version (within); the first versus second fixation-count half of each reading (within); high versus low or no self-rated programming expertise, excluding ``medium'' (between); correct versus incorrect answer, pooled over both programs (mixed); and the same restricted to the Rectangle program and to the Vehicle program (between, one trial per participant). Pooled over the programs, the correctness label is confounded with the program: among the retained trials, 92 of 133 Rectangle answers were correct and 106 of 138 Vehicle answers incorrect. We report that problem deliberately and compare it with the two within-program versions. Line spacing of the code is 37 px.

\paragraph{GazeBase.} GazeBase \citep{griffith2021gazebase} contains 1000 Hz monocular EyeLink recordings in degrees of visual angle on a $1680\times1050$ px display at 550 mm ($\pm 23.3^\circ\times\pm 15.1^\circ$). We took all 322 participants of Round~1, both sessions, and the tasks reading (TEX), video 1 and 2 (VD1, VD2; consecutive 60-s segments of one film trailer) and the gaze-driven game (BLG). Fixations were detected with I-VT on the raw samples (5-sample moving average; $30^\circ/\mathrm{s}$; minimum 80 ms) for every task, because the dataset's own fixation labels are absent for the video and game tasks. Five problems were formed: TEX versus VD1, VD1 versus BLG, VD1 versus VD2, the first versus second fixation-count half of each VD1 recording, and reading in session 1 versus session 2 (different passages). There are no text lines in most of these tasks, so the reference scale is $2^\circ$.

\paragraph{A common spatial scale.} Every spatial constant is expressed in units of the reference scale $u$ (one text line in SB-SAT and EMIP; $2^\circ$ in GazeBase): recurrence radius $0.5u$ and $1u$, directed-graph tile $1u\times1u$, jitter $0.25u$, $0.5u$, $u$ and $2u$, and lattice anchors $2u$ apart. A sensitivity to jitter at $0.25u$ in SB-SAT is therefore a sensitivity to structure finer than a quarter of a line, about one word in height.

\begin{table}[!htbp]
\centering\small
\caption{The thirteen problems. Label level: whether the label varies within participants, is constant within each participant, or both. Trials and participants after exclusions.}
\label{tab:problems}
\resizebox{\linewidth}{!}{\input{tab_problems}}
\end{table}

\subsection{Descriptors}\label{sec:desc}
Nine descriptors were profiled (ten variants, counting the two recurrence radii): two baselines that read one property by construction, two descriptors of durations and amplitudes, recurrence, and four graph descriptors.

\paragraph{Count (baseline).} The number of fixations. It is invariant to every transformation, so its profile is identically zero and its AUC measures how much of a problem is count.

\paragraph{Dispersion (baseline).} The SD of $x$, the SD of $y$, and the mean pairwise distance between fixations. All three are translation-invariant: the baseline reads the extent and shape of the spread, not where on the screen it lies. It is invariant to order and to both duration transformations; the two SDs read extent along the screen axes and therefore change under rotation when the spread is elongated, whereas the mean pairwise distance is rotation-invariant.

\paragraph{Classic.} Mean fixation duration and mean saccade amplitude (Euclidean distance between consecutive fixations).

\paragraph{Duration trend.} The correlation of fixation duration with sequence position, and the mean duration. It reads how durations evolve over a trial, so it is sensitive to order and to both duration transformations, and it serves as a positive control for the two duration transformations on real data.

\paragraph{Recurrence quantification} \citep{anderson2013recurrence}: recurrence rate, determinism, laminarity, and centre of recurrence mass, at radius $0.5u$ and $1u$. Recurrence rate is order-invariant; determinism and laminarity, which count diagonal and vertical line structures in the recurrence plot, are not.

\paragraph{Directed transition graph.} Fixations are assigned to $1u\times1u$ tiles, consecutive repeats are collapsed, and a directed edge $i\to j$ counts immediate transitions; four features summarize the matrix: edge reciprocity, revisit tendency ($1 - |V|/|s|$), directed transition entropy, and degree asymmetry. Tiles are unbounded, so the descriptor remains defined after rotation.

\begin{figure}[!htbp]
\centering
\includegraphics[width=\linewidth]{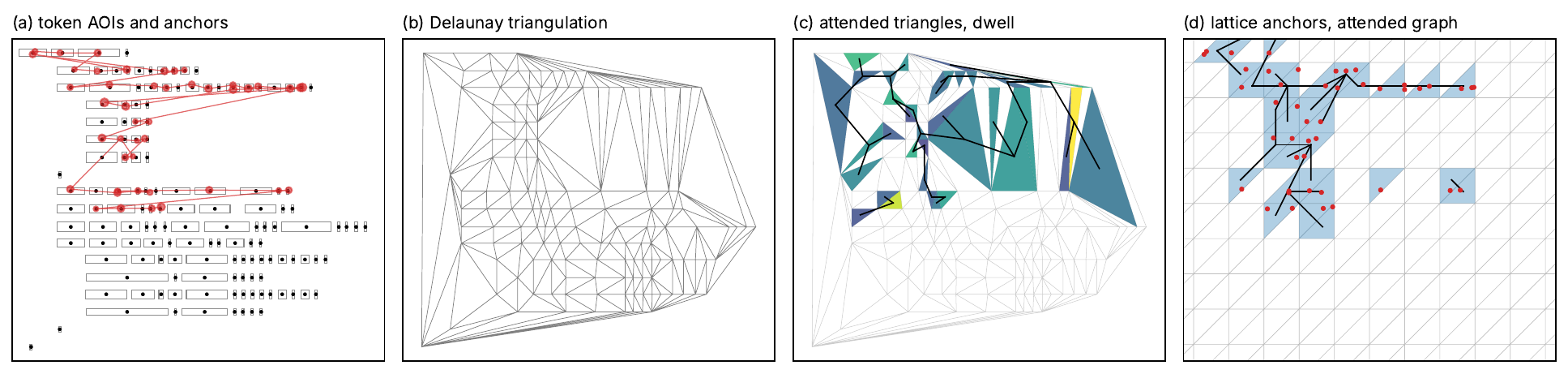}
\caption{The Fixation Adjacency Graph on a \emph{synthetic} trial (red; marker size is duration) over the token areas of interest of the EMIP Rectangle program; no participant's gaze and no stimulus image are shown. (a) Token AOIs and their centres; (b) the Delaunay triangulation of the centres; (c) attended triangles coloured by dwell, with the maximum spanning tree of the graph (black); (d) the same trial on the stimulus-independent lattice graph, anchors $2u$ apart.}
\label{fig:fag}
\end{figure}

\paragraph{Fixation Adjacency Graph.} The undirected graph treats a trial's gaze as a graph on regions of the stimulus (Figure~\ref{fig:fag}). Anchor points are joined by a Delaunay triangulation \citep{lee1980two}; each triangle accumulates the dwell of the fixations inside it; attended triangles are the vertices, two are adjacent when they share a vertex of the triangulation, and edges are weighted by combined dwell. Five features summarize the graph: attention centrality (the weight of the largest component's maximum spanning tree relative to the total), the maximum edge betweenness \citep{brandes2001faster}, a compactness index based on the Estrada index \citep{estrada2000characterization}, a squareness index of each component's extent, and the number of components. The graph was introduced on segmentations of diagram stimuli \citep{mozaffari2020classification, mozaffari2021quantifying}; here we build it on two kinds of anchor that any stimulus provides. In the \emph{AOI graph} (EMIP only) the anchors are the centres of the code tokens' areas of interest, so the graph lives on the program's own lexical layout. In the \emph{lattice graph} (every problem) the anchors are the nodes of a square lattice $2u$ apart, which covers the disc swept by the display under rotation, so the rotation transformation never moves a fixation off the support. The AOI graph's support is bounded; it is rotated about the centre of its anchors, and the fixations a transformation moves off the support are dropped. Because that loss can differ between classes, we also profile a \emph{padded AOI graph}, in which lattice anchors $2u$ apart surround the token anchors out to the disc swept by the display under rotation (anchors closer than one spacing to a token anchor are left out), so that the triangulation inside the code is essentially unchanged and no transformation moves a fixation off the support (Section~\ref{sec:support}).

\input{sec_results}

\section{Discussion}
\subsection{What the Profile Adds}
A classification accuracy says that a descriptor works; a perturbation profile says how much of that discrimination survives when each property of the scanpath is destroyed and the pipeline is refit. On thirteen problems from three public datasets it did three things an accuracy cannot. It explained a counter-intuitive result, a descriptor that discriminates \emph{better} without the information its name says it uses, and the Proposition of Section~\ref{sec:transf} turns that explanation into an exact statement (Section~\ref{sec:against}). Together with a within-program permutation scheme, a same-program AUC and a synthetic-gaze control, it showed that a ``comprehension'' contrast, and the program contrast beneath it, were contrasts between stimulus layouts (Section~\ref{sec:confound}). And it located, for each pipeline, the property and the spatial scale its discrimination depends on: durations and their evolution for the tasks and segments of GazeBase, structure coarser than about $2^\circ$ for early versus late viewing, and structure at the scale of one to two text lines for recurrence on first- versus second-language reading (Sections~\ref{sec:native} and~\ref{sec:gb}). In each case the baselines were essential. Count and dispersion read one property by construction, and comparing a structured descriptor's AUC and profile with theirs is what turns a sensitivity into an interpretation.

The checks around the profile mattered as much as the profile. A stated selection rule kept 6 of 95 intervals that excluded zero out of the interpretation; the pooled metric, the fold partitions and the longer runs removed 7 more; the padded graph removed the two sensitivities of the AOI graph, which were partly differential loss of support; and a second learner confirmed 45 of the remaining 80. The analysis also returned a negative result about one of our own descriptors. On these data the Fixation Adjacency Graph on stimulus anchors supported no claim that a simpler descriptor or the layout itself did not explain (Section~\ref{sec:graph}).

\subsection{Practical Guidance}\label{sec:guide}
The results translate into a short procedure for anyone profiling a descriptor.
\begin{enumerate}
\item \emph{Profile two baselines alongside every descriptor.} Fixation count and spatial dispersion each read one property by construction. A structured descriptor whose AUC the dispersion baseline matches, or whose profile mirrors the baseline's, is not shown to use anything beyond spread (Sections~\ref{sec:against} and~\ref{sec:confound}). Where stimuli differ between classes, add a stimulus-identity baseline.
\item \emph{Use a within-fold AUC with class-balanced folds, and read every AUC against its permutation distribution.} Under a between-participant label the pooled cross-validated AUC sits below 0.5 when there is no signal (0.43 on average in the simulations with unbalanced folds), and perturbations can change this bias. Permute labels in the way the design assigned them; where a stimulus is associated with the label, permute within stimuli, and expect the permutation mean to move away from 0.5 (Sections~\ref{sec:fold} and~\ref{sec:confound}).
\item \emph{State the rule by which cells are interpreted, and fix it before the final analysis.} Require that the observed AUC exceeds its null and that the sensitivity interval excludes zero, then check each such cell against a second metric, several fold partitions and a longer run, and report all of them in a ledger (Section~\ref{sec:ledger}).
\item \emph{Look for a mechanism behind a negative sensitivity before reading it.} A transformation can make information accessible that the observed features express poorly, as order permutation does for saccade amplitude, or it can merely change an evaluation's bias; the three permutation distributions of Section~\ref{sec:stat} tell these apart. Of the 19 robust negative sensitivities here, only the two with a known mechanism held under a second learner (Sections~\ref{sec:against} and~\ref{sec:learners}).
\item \emph{Treat a rotation sensitivity of a spread descriptor as a warning about stimulus layout.} Then compute the AUC over pairs of trials that share a stimulus, and profile synthetic gaze drawn from each stimulus's layout: a descriptor that separates the synthetic trials is reading the stimulus (Section~\ref{sec:confound}).
\item \emph{For a descriptor with bounded support, report what the transformations leave on the support.} Report retention by transformation and class, profile a retention-only descriptor, and repeat the cells you interpret on a padded version of the descriptor that loses nothing (Section~\ref{sec:graph}).
\item \emph{Do not read a zero sensitivity as evidence that a property is unused.} After refitting, a learner can move to redundant information, and a wide interval covers zero for lack of precision. Use an explicit equivalence margin, and treat exact invariances as checks of the implementation, not results.
\item \emph{Start with a linear learner and repeat the cells you interpret with a nonlinear one.} Count a sensitivity as holding under both only if both intervals exclude zero; a kept sign is weaker evidence. Expect small negative sensitivities at ceiling to be specific to the linear learner (Section~\ref{sec:learners}).
\item \emph{Plan for conservative intervals, and for the sample size they require.} Under no signal, the mean width was 0.12 at 95 participants $\times$ 20 trials, about 0.35 at 60 $\times$ 4 and 112 $\times$ 2, and 0.46 at 65 $\times$ 2 (Table~\ref{tab:cal}). At these sizes, simulated sensitivities of 0.12--0.14 had intervals excluding zero in only 8--60\% of datasets, while sensitivities of 0.30 did in all. With 322 participants the mean width on the real problems was 0.05, and nine tenths of all robust cells came from that dataset.
\item \emph{Use modest Monte Carlo settings for screening and larger ones for the cells you report.} $R = 6$ and $B = 60$ gave conservative intervals in the simulations and reached the same conclusion as $R = 20$ and $B = 300$ under three seeds in 82 of 89 cells; the exceptions were cells whose intervals ended near zero (Section~\ref{sec:conv}). Report the seeds.
\item \emph{State spatial scales in a unit the stimulus defines.} Expressing radii, tiles and jitter in line spacings made a quarter-line or two-line sensitivity comparable across datasets with different displays.
\end{enumerate}

\subsection{Relation to Prior Work}
Perturbing scanpaths to learn what a measure responds to is an established paradigm. \citet{dewhurst2012multimatch} validated MultiMatch with synthetic spatial-noise and order manipulations, and \citet{fahimi2021metrics} applied a systematic family of perturbations, including spatial noise, ordinal shifts and reversals, to a catalogue of scanpath \emph{similarity} metrics in order to characterize their robustness; our jitter transformation is operationally close to their spatial noise, and their ordinal manipulations are specific order disruptions. Shuffled-order baselines are standard in recurrence analysis \citep{anderson2013recurrence, anderson2015comparison}, static-versus-sequential comparisons appear in task decoding \citep{greene2012reconsidering}, and \citet{meghanathan2026temporal} used ordered versus shuffled gaze sequences directly. In the broader literature on predictive importance, retraining after removing information follows \citet{hooker2019benchmark}, and \citet{williamson2023general} give a general framework for the importance of a variable group as the loss in population predictiveness when the model is refit without it, with asymptotically valid inference. A sensitivity $\Cs_\tau$ is of that kind, with a transformed scanpath taking the place of a removed feature.

What the profile adds operationally is the combination, specialized to scanpath decoding: (i) transformations defined on whole scanpaths rather than on features, each with a stated list of invariants, so that the transformed data remain scanpaths and the properties a transformation leaves intact are known; (ii) exact invariance checks that follow from those lists and catch implementation errors; (iii) participant-aware evaluation, with participant-held-out folds, permutation schemes that respect how labels were assigned \citep{ojala2010permutation}, and a within-fold AUC that guards against evaluation bias; (iv) a refit participant bootstrap whose coverage was checked by simulation, including weak, negative and unbalanced designs, and found conservative; and (v) a reusable package with the baselines and support diagnostics described above. We do not claim priority for any individual transformation, and unlike \citet{williamson2023general} we offer no asymptotic theory for the interval; its support is the simulation evidence of Section~\ref{sec:sim}.

\subsection{Limitations}
The transformations are not orthogonal and do not partition the signal: jitter alters saccade amplitudes as well as recurrence, and the order transformation preserves recurrence rate but not determinism or laminarity. Rotation preserves eccentricity and any rotationally invariant location contrast. The jitter results are properties of the sweep grid, not spatial scales of behaviour. The framework is binary. The intervals are unadjusted and exploratory. Their coverage was checked on one generative family, mostly with 40 datasets per cell, and at 322 participants only for the no-signal and signal-elsewhere designs; per-contrast error rates are bounded above by 7--15\% rather than shown to be 5\%. Pooled over contrasts, the intervals were conservative in every design tested (single contrasts covered in 92.5\% to 100\% of 40 datasets), with standard errors nearly twice the empirical spread, so the analysis has less power than its nominal level suggests and an interval that covers zero is weak evidence of absence; we did not attempt to recalibrate them. The behaviour of the ``negligible'' call near its margin was not re-examined. The selection rule was fixed before this version of the analysis but after earlier versions of it had been seen, so it is a guard against selective reading, not a preregistration. The duration donor pool is built once on the other participants of the full dataset, not within training folds; a training-fold pool changed the interpreted duration sensitivities by at most 0.007. With 322 participants, sensitivities of a few thousandths of an AUC are resolved, and statistical robustness there should not be mistaken for practical importance. The second learner was run at the screening settings, and at the long settings only for the 21 cells of Table~\ref{tab:robust}. The problems are demonstrations: they discriminate specific recordings, groups and stimuli, and license no general claim about reading, programming or video viewing. Between-participant labels (native language, expertise) cannot separate the named factor from any other difference between the groups. The EMIP release analysed keeps only fixations on code tokens, so every EMIP profile describes on-code looking, and expertise in EMIP is self-rated. Finally, a profile characterizes a descriptor, learner and evaluation on a task; it does not by itself say what participants were doing.

\section{Conclusions}
Before choosing a scanpath descriptor for a classification problem, one can ask how much of its discrimination survives when each property of the scanpath is destroyed and the classifier is refit. Five transformations with stated invariants answer that question for any descriptor, with built-in invariance checks. On three public datasets the answer explained exactly why order permutation can improve a classifier, showed that a comprehension contrast was a contrast between stimulus layouts, and identified, among 565 intervals, the 80 sensitivities that survive a change of evaluation metric, of fold partition and of Monte Carlo settings, and the 45 that also survive a change of learner. The tool is released so that the question, and these checks, can be applied routinely.

\section*{Data Availability}
All data analysed are public: SB-SAT at \url{https://github.com/ahnchive/SB-SAT}, EMIP at \url{https://osf.io/53kts} with its corrected release at \url{https://github.com/nalmadi/EMIP-Toolkit}, and GazeBase at its figshare record \citep{griffith2021gazebase}. GazeBase and EMIP are distributed under the Creative Commons Attribution 4.0 licence; SB-SAT and the corrected EMIP release state no licence and are used as published, with the citations their authors request \citep{ahn2020towards, almadi2021emiptoolkit}. No raw recording is redistributed; the repository contains only a fixation table derived from GazeBase, shared under the same CC BY 4.0 licence. The package \pkg\ (MIT licence), the scripts that download and preprocess each dataset, every analysis script, every aggregate output and the manuscript source are available at \url{https://github.com/saleh-mozafari/scanpath-perturbation-profiles}, archived on Zenodo at \url{https://doi.org/10.5281/zenodo.24132704}.

\section*{Ethics Statement}
This study is a secondary analysis of publicly available, de-identified eye-tracking datasets, each collected under its authors' own ethics and consent procedures as described in the cited publications. No new data were collected and no participant was contacted. Formal ethics approval was therefore not required.

\section*{Author Contributions}
Conceptualization, S.M.\ and A.N.; methodology, S.M.; software, S.M.; formal analysis, S.M.; writing---original draft preparation, S.M.; writing---review and editing, A.N.; supervision, A.N.

\section*{Funding}
This research received no external funding.

\section*{Acknowledgments}
The authors thank the creators of SB-SAT, EMIP and its corrected release, and GazeBase for making their data public. The authors used Claude (Anthropic), a large language model, to edit language, to draft and revise analysis code and manuscript text under the authors' direction, and to check reported statistics against stored outputs. The authors reviewed and tested the code, verified the analyses and the text, and take full responsibility for the content of this publication.

\section*{Conflicts of Interest}
The authors declare no conflicts of interest.

\clearpage
\appendix
\section{Supplementary Tables}\label{app:supp}
\setcounter{table}{0}\renewcommand{\thetable}{S\arabic{table}}
\providecommand{\Cs}{C}
\begin{landscape}
Tables S1--S13: every profile of the primary analysis (within-fold AUC, class-balanced participant folds); $R = 6$ replicates, $B = 60$ refit-bootstrap draws, seed 42; linear SVM, ten-fold participant-held-out cross-validation; duration-marginal donor pool of the other participants' durations. $u$ is the reference scale (one text line in SB-SAT and EMIP, $2^\circ$ in GazeBase). The same profiles with the pooled AUC are in the released outputs.
\input{supp_profiles}

\begin{table}[!htbp]\centering\scriptsize
\caption{Observed pooled AUC (all out-of-fold decision values ranked together; class-balanced participant folds), with the permutation $p$-value in brackets (200 permutations under the schemes of Table~2).}
\resizebox{\linewidth}{!}{\input{tab_auc_pooled}}
\end{table}

\begin{table}[!htbp]\centering\scriptsize
\caption{Range, over descriptors, of the mean of the permutation distribution of the observed AUC, under the within-fold AUC (1000 permutations) and the pooled AUC (200 permutations), both with class-balanced folds. On \texttt{em\_correct} labels are permuted within programs and on \texttt{sb\_compr} between the passages of a participant, so stimulus structure that survives the permutation raises the mean for descriptors that read it.}
\input{supp_null_means}
\end{table}

{\scriptsize\setlength{\tabcolsep}{2.5pt}\setlength{\LTcapwidth}{\linewidth}
\input{supp_ledger}
}

\begin{table}[!htbp]\centering\scriptsize
\caption{Bootstrap standard error against the empirical spread of the estimate in the calibration campaign ($R = 6$, $B = 60$; 40 datasets per design and size). SE ratio: mean bootstrap SE divided by the SD of the estimate across datasets, per contrast; median and range over the contrasts of the cell. Coverage of the target by normal intervals at nominal 80\%, 90\% and 95\%.}
\input{supp_se}
\end{table}

\begin{table}[!htbp]\centering\scriptsize
\caption{Corrected refit bootstrap (Equation~\ref{eq:var}) against the direct refit bootstrap, which averages all $R = 6$ transformed replicates inside each of the $B = 60$ draws, on the same 200 datasets per cell. Coverage of the target, mean width, and the share of intervals excluding zero.}
\input{supp_direct}
\end{table}

{\scriptsize\setlength{\LTcapwidth}{\linewidth}
\input{supp_ext}
}

\begin{table}[!htbp]\centering\scriptsize
\caption{Controls for the EMIP program and correctness contrasts (within-fold AUC). Program contrast: AUC on real gaze and on synthetic gaze, in which every trial is replaced by the same number of fixations drawn uniformly over the bounding box of its program's code tokens. Correctness, both programs: AUC over all correct--incorrect pairs of a test fold, the mean of its permutation distribution when labels are permuted within programs, and the AUC over pairs from the same program and test fold with a refit participant-bootstrap interval (200 draws) and a one-sided permutation $p$-value (1000 permutations within programs). Descriptors with no spatial features are not affected by the synthetic control and are marked with a dash.}
\input{supp_controls}
\end{table}

\begin{table}[!htbp]\centering\scriptsize
\caption{EMIP expertise: AUC of each of the ten test folds and of the corresponding training set, for three descriptors (linear SVM; class-balanced folds).}
\input{supp_expertise}
\end{table}

\begin{table}[!htbp]\centering\scriptsize
\caption{Share of dwell time on the AOI triangulation, negative / positive class, for every EMIP problem and transformation (mean over trials of the mean over $R = 6$ replicates). Fixations off the triangulation are dropped by the AOI graph. On the lattice graph and on the padded AOI graph the share was at least 0.998 in every cell.}
\resizebox{\linewidth}{!}{\input{supp_retention}}
\end{table}

\begin{table}[!htbp]\centering\scriptsize
\caption{AOI-graph sensitivities on the two halves of a program reading under the default transformations and under support-matched transformations (redrawn until the trial keeps at least its observed dwell share minus 0.02; in brackets, the share of draws for which this succeeded within 200 tries), and the sensitivities of a descriptor made only of the fixation and dwell shares on the support. In the first column: observed AUC of the AOI graph and of the support-only descriptor. Bold: interval excludes zero.}
\resizebox{\linewidth}{!}{\input{supp_matched}}
\end{table}

\begin{table}[!htbp]\centering\scriptsize
\caption{Duration-marginal transformation of the classic descriptor with the leave-one-participant-out donor pool of the main analysis and with training-fold pools (each fold's trials draw only from that fold's training participants; mean and SD over $R = 6$ replicates; within-fold AUC, class-balanced folds).}
\input{supp_donor}
\end{table}

\begin{table}[!htbp]\centering\scriptsize
\caption{Three permutation distributions for negative sensitivities (within-fold AUC, class-balanced folds; 1000 permutations under the schemes of Table~2; $R = 6$ transformed replicates). The observed AUC, the transformed AUC and their contrast are each compared with their own permutation distribution; for the contrast the lower-tail probability is given. The first four rows are the order cells of the classic descriptor discussed in Section 5.4 of the article; the last is the unexplained cell of Section 5.11.}
\input{supp_nulls}
\end{table}

\begin{table}[!htbp]\centering\scriptsize
\caption{Calibration at the largest analysed sample size: 322 participants $\times$ 4 trials with the label varying within participants ($R = 6$, $B = 60$; 40 datasets per row; targets from 60 further datasets with $R = 20$, or zero by exchangeability for the order contrast). Coverage of the target by the 95\% interval; exclusions of zero where the target is zero by argument; mean width; mean bootstrap SE over the empirical SD of the estimate; coverage of the nominal 80\% interval.}
\input{supp_big}
\end{table}

\begin{table}[!htbp]\centering\scriptsize
\caption{Gradient-boosted trees on the cells of Table~7 at the screening settings and at the long settings (seed 42). Bold: interval excludes zero.}
\input{supp_trees_long}
\end{table}

\clearpage
\section*{Methods appendix}
\subsection*{A1. Feature definitions}
Let a trial have fixations $(x_i, y_i, d_i)$, $i = 1, \dots, N$, in sequence order, with $d_i$ the duration.

\emph{Count.} $N$.

\emph{Dispersion.} The population SDs of $x$ and of $y$, and the mean pairwise distance $\binom{N}{2}^{-1} \sum_{i<j} \lVert p_i - p_j \rVert$, $p_i = (x_i, y_i)$.

\emph{Classic.} Mean duration $N^{-1} \sum_i d_i$ and mean saccade amplitude $(N - 1)^{-1} \sum_{i<N} \lVert p_{i+1} - p_i \rVert$.

\emph{Duration trend.} The Pearson correlation of $d_i$ with $i$, and the mean duration; trials with fewer than five fixations are excluded.

\emph{Recurrence} \citep{anderson2013recurrence}. Fixations $i < j$ recur if $\lVert p_i - p_j \rVert < r$; $n_r$ is the number of recurrent pairs, and the recurrence plot is the upper triangle of the $N \times N$ recurrence matrix. Recurrence rate: $100 \cdot 2 n_r / (N(N-1))$. Determinism: $100$ times the share of recurrent pairs that lie on diagonal lines of length at least 2. Laminarity: $100$ times the share that lie on vertical lines of length at least 2. Centre of recurrence mass: $100 \cdot \overline{(j - i)} / (N - 1)$, the mean over recurrent pairs. A trial without recurrent pairs has all four equal to zero; trials with fewer than five fixations are excluded. Radii: $r = 0.5u$ and $1u$.

\emph{Directed transition graph.} Each fixation is assigned to the tile $(\lfloor x/u \rfloor, \lfloor y/u \rfloor)$; consecutive repeats are collapsed to a sequence $s$ over the visited tiles $V$; $A_{ab}$ counts immediate transitions $a \to b$, and $E = \{(a, b): a \neq b, A_{ab} > 0\}$. Reciprocity: the share of edges in $E$ whose reverse is in $E$. Revisit tendency: $1 - |V|/|s|$. Transition entropy: $-\sum_{ab} P_{ab} \log_2 P_{ab}$ with $P = A / \sum A$. Degree asymmetry: the mean over tiles of $|\mathrm{out}_a - \mathrm{in}_a| / (\mathrm{out}_a + \mathrm{in}_a)$, with out- and in-strengths the row and column sums of $A$.

\emph{Fixation Adjacency Graph.} Anchors are triangulated (Delaunay). The dwell of a triangle is the summed duration of the fixations inside it; fixations in no triangle are dropped. Vertices are the triangles with positive dwell; two are adjacent if they share an anchor; the weight of an edge is the summed dwell of its two triangles. Attention centrality: the largest, over connected components, of the weight of the component's maximum spanning tree, divided by the total edge weight (set to 0.25 for a graph without edges). Betweenness: the maximum edge betweenness centrality \citep{brandes2001faster}. Compactness: the Estrada index $\sum_k e^{\lambda_k}$ of the adjacency matrix \citep{estrada2000characterization}, averaged over components. Squareness: for each component, $\Delta_x \Delta_y / \max(\Delta_x, \Delta_y)^2$ with $\Delta_x, \Delta_y$ the extents of its triangle centroids (zero for a single triangle), averaged over components. The fifth feature is the number of components. Lattice anchors are $2u$ apart and cover the disc swept by the display under rotation; AOI anchors are the token centres of the program; the padded AOI graph adds lattice anchors $2u$ apart around them, leaving out those closer than $2u$ to a token anchor.

\subsection*{A2. Class sizes and fixation counts}
\begin{table}[H]\centering\scriptsize
\caption{Trials and participants per class, and fixations per trial, for every problem (after the exclusions of Section 4). Class labels are those of the released loaders.}
\input{supp_descriptives}
\end{table}

\subsection*{A3. Software and settings}
Python 3.13.16; NumPy 2.5.3; SciPy 1.18.1; pandas 3.0.5; scikit-learn 1.7.2; XGBoost 3.2.0; NetworkX 3.7; package \texttt{scanpathdecomp} 0.6.3. Linear SVM: \texttt{LinearSVC} (liblinear; squared hinge loss, L2 penalty, $C = 1$, at most 20\,000 iterations, random state 0) on features standardized with training-fold means and SDs. Gradient-boosted trees: \texttt{XGBClassifier} with 200 trees of depth 3, learning rate 0.05, row subsampling 0.8, histogram tree method, logistic loss, random state 0. Folds: ten participant folds, participants dealt within class in a seeded order (seed 0 for the reported profiles; seeds 0--19 for the partition check).
\end{landscape}

\bibliographystyle{apalike}
\bibliography{references}
\end{document}

%% file: sec_intro_a.tex
Trial-level scanpath descriptors are the workhorse of gaze-based classification \citep{selim2024review}. Recurrence quantification analysis summarizes re-inspection \citep{anderson2013recurrence}; transition matrices and their graph-theoretic reductions summarize the ordered structure of movement between regions \citep{smith2010eye, popelka2016scangraph}; entropy measures summarize the dispersion and predictability of looking; and mean fixation duration and saccade amplitude remain in nearly every feature set \citep{holmqvist2011eye}. When one of these descriptors discriminates a task \citep{greene2012reconsidering, borji2014defending, coutrot2018scanpath, shubi2025decoding}, an expertise level \citep{gegenfurtner2011expertise}, or a cognitive state, the result is usually reported as an accuracy. What is rarely reported is which properties of the scanpath the discrimination rests on: the set of locations, their order, how long each was held, or where on the stimulus they fell. Different descriptors encode these properties in different mixtures, and a descriptor's name is a poor guide to what it uses in a given dataset.

The consequences are concrete. In one of the public datasets analysed below, a classifier separates programmers who answered a comprehension question correctly from those who did not with an AUC of 0.72, using only the extent and shape of their fixations' spread. Most wrong answers came after one of the two programs, whose code has a different shape on the screen; when the answers are shuffled within each program the same classifier still averages 0.71, and among trials of the same program the spread carries no detectable signal. An accuracy cannot reveal this. Asking how the discrimination changes when a stated property of every scanpath is destroyed and the classifier is retrained, and comparing the answer with simple baselines, can.

Perturbing scanpaths to find out what a measure is sensitive to has precedent. \citet{dewhurst2012multimatch} validated MultiMatch with synthetic spatial-noise and order manipulations, and \citet{fahimi2021metrics} formalized this into a family of perturbations applied across a catalogue of scanpath similarity metrics to characterize their robustness. Shuffled-order baselines are standard in recurrence analysis \citep{anderson2013recurrence, anderson2015comparison}, static-versus-sequential comparisons appear in task decoding \citep{greene2012reconsidering}, and ordered-versus-shuffled gaze sequences have been used to dissociate cortical representations \citep{meghanathan2026temporal}. Outside eye tracking, retraining after removing information is a standard way to measure predictive importance \citep{hooker2019benchmark}, with formal inference for the resulting loss in predictiveness \citep{williamson2023general}. The scanpath studies ask how robust a \emph{similarity metric} is, or whether order matters for one task, and the importance methods operate on features rather than on scanpaths. Neither provides a general procedure for asking which properties of a scanpath a \emph{classification pipeline} depends on, with transformations whose invariants are stated, participant-aware evaluation, and uncertainty that has been checked.

This paper provides that procedure, the \emph{perturbation profile}, and makes four contributions. (1) Five transformations, each targeting one property of a scanpath (fixation order, the assignment of durations to fixations, the duration distribution, orientation and handedness, and fine spatial structure) while preserving a stated list of invariants, including the properties a transformation's name suggests it removes but does not. (2) A refit participant bootstrap for the change in participant-held-out AUC, whose coverage we check by simulation at the sample sizes analysed, together with exact invariance checks that catch implementation errors, a within-fold AUC that guards against evaluation bias, support diagnostics for descriptors defined on a bounded region, and a stated rule for which sensitivities are interpreted. (3) Demonstrations on thirteen problems from three public datasets (reading, program comprehension, and video and game viewing) in which the profile explains, with an exact result, why a descriptor improves when its own information is destroyed, helps identify a stimulus confound, and shows which sensitivities survive a change of evaluation metric, fold partition, Monte Carlo settings and learner. (4) An open Python package that profiles any descriptor with any scikit-learn learner. Section~\ref{sec:guide} turns the results into practical guidance.

The quantity measured throughout is the sensitivity of a refitted pipeline (descriptor, learner and participant-held-out evaluation together) to the destruction of a stated property. It is not an attribution of what a fitted classifier relied on, and it is not a measure of what participants did; the datasets below are demonstrations of that quantity rather than tests of hypotheses about viewing behaviour. The profile is meant for anyone who reports a scanpath-based classification and wants to say what the result rests on, and for anyone choosing between descriptors.

%% file: sec_method.tex
\section{The Perturbation Profile}
\subsection{Setting}
A trial is a time-ordered sequence of fixations $\{(x_k, y_k, d_k)\}_{k=1}^{n}$ with locations in a common frame and durations $d_k$. A descriptor $f$ maps a trial to a feature vector. A task supplies a binary label and a participant identifier per trial. Performance is the AUC of out-of-fold decision values from participant-held-out cross-validation: participants are split into ten folds and all trials of a participant are held out together, so no participant contributes to both training and test, and features are standardized on the training folds only. A cross-validated AUC can be computed in two ways. The \emph{pooled} AUC ranks all out-of-fold decision values together. Pooling compares scores produced by different fitted models, and differences in their score offsets and in the class proportions of their training folds can bias it; with little or no signal, the bias is usually downward \citep{parker2007stratification, airola2011comparison}, most of all when the label is constant within participants and the folds differ in their class mix. A perturbation can change this bias as well as the discrimination, so a difference between observed and transformed pooled AUCs can contain both. The primary analysis therefore uses the \emph{within-fold} AUC, the AUC over positive--negative pairs drawn from the same test fold (equivalently, the per-fold AUCs averaged with weights proportional to their numbers of pairs), with \emph{class-balanced} folds: participants are dealt to the folds within each class, so that every fold receives a similar mix of classes. Within a fold all scores come from one model, so offsets between models cannot enter. Every profile is repeated with the pooled AUC as a sensitivity analysis. Cross-validation estimates the performance of the fitting procedure over training sets rather than that of one fitted model \citep{bates2024cross}, which is the quantity a pipeline-level sensitivity requires. A within-participant AUC would remove participant offsets too, but it answers a different question and is undefined for the five problems whose label is constant within participants. The learner is a linear support vector machine (liblinear, $C = 1$) on standardized features; the framework accepts any scikit-learn estimator.

\subsection{Transformations}\label{sec:transf}
\begin{figure}[!htbp]
\centering
\begin{adjustwidth}{-\extralength}{0cm}
\centering
\includegraphics[width=\linewidth]{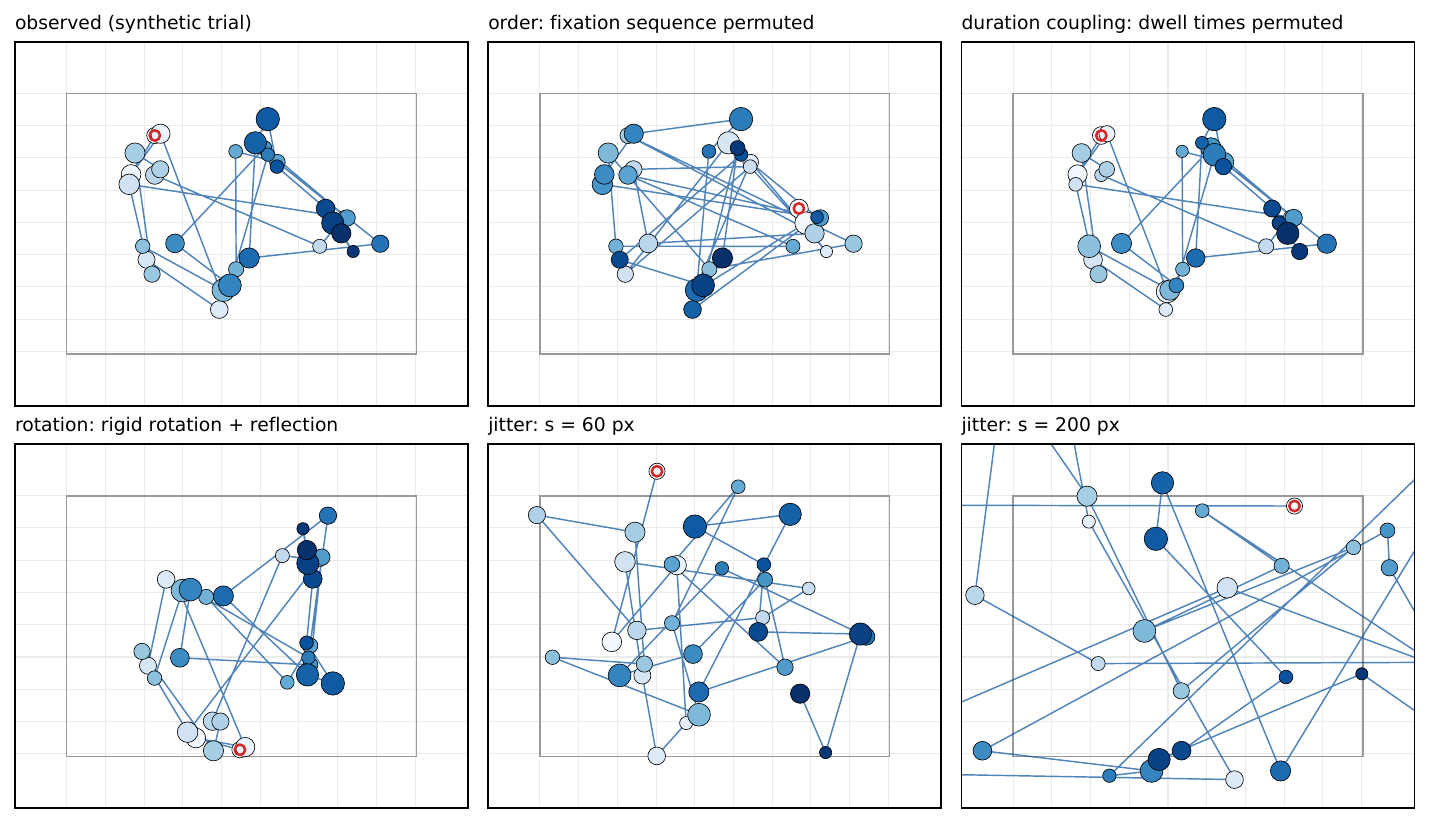}
\end{adjustwidth}
\caption{One \emph{synthetic} trial in the analysis window under each transformation; no participant's gaze is shown. Marker size is fixation duration, colour is order (light to dark), the red ring is the first fixation, the light grid is the tiling of the directed graph (Section~\ref{sec:desc}). Shown are four of the five transformations (duration marginal would look like duration coupling with different marker sizes). Order permutation keeps the points and re-threads the path; duration coupling keeps the path and re-sizes the markers; rotation keeps every distance, angle and eccentricity and changes orientation and side; jitter at 60 px perturbs fine spatial structure, at 200 px coarse structure.}
\label{fig:schematic}
\end{figure}

Each transformation $\tau$ is applied independently to every trial (Table~\ref{tab:nulls}; Figure~\ref{fig:schematic}). Each targets a stated property while preserving an explicitly listed set of invariants, and interpretation depends on that list: a sensitivity to $\tau$ is a sensitivity to whatever $\tau$ changes, which is always the named property together with anything that depends on it.

\begin{table}[!htbp]
\centering\small
\caption{The five transformations, with what each destroys and what each preserves. For the first four the preservation statements are exact; jitter is a gradual perturbation, and its row describes distribution-level effects. The last column names the property a transformation does \emph{not} remove although its name might suggest it.}
\label{tab:nulls}
\begin{adjustwidth}{-\extralength}{0cm}
\centering
\begin{tabular}{p{2.1cm}p{3.4cm}p{2.5cm}p{3.5cm}p{2.5cm}}
\toprule
Transf. & Operation & Destroys & Preserves & Not removed \\
\midrule
order & permute the fixation sequence & serial order & multisets of locations and durations; every order-free property & --- \\
duration coupling & permute durations across the trial's fixations & which fixation carries which duration, and with it any dependence of duration on sequence position (trends, serial autocorrelation) & locations, order, the trial's duration distribution (mean, spread, total dwell, count) & marginal duration information \\
duration marginal & redraw the trial's durations from the pooled durations of the other participants & the trial's duration distribution, the coupling and the duration--sequence dependence & locations, order, count & --- \\
rotation & rigid rotation (uniform angle) and random reflection about a centre & orientation, side relative to the stimulus, and handedness (chirality) & all pairwise distances and unsigned angles, order, durations, each fixation's eccentricity from the centre & eccentricity; any rotationally invariant location contrast (exact) \\
jitter($s$) & add isotropic Gaussian noise, SD $s$, to every location & fine spatial structure, increasingly with $s$ (shorthand; not a band-pass filter) & order and durations (exact); orientation and structure coarser than $s$ approximately, at the distribution level & --- \\
\bottomrule
\end{tabular}
\end{adjustwidth}
\end{table}

The \emph{order} transformation removes only the sequence. It can also change what an order-dependent feature measures, and for one common feature the change is exact. Let a trial have fixations at $p_1,\dots,p_n$ ($n \geq 2$), let $A = (n-1)^{-1}\sum_{k=1}^{n-1}\lVert p_{k+1} - p_k\rVert$ be its mean saccade amplitude, and let $\bar D = [n(n-1)]^{-1}\sum_{i \neq j}\lVert p_i - p_j\rVert$ be its mean pairwise distance. \emph{Proposition.} If $\pi$ is a uniformly random permutation of $\{1,\dots,n\}$ and $A_\pi$ the mean amplitude of the permuted sequence, then $\mathbb{E}_\pi[A_\pi] = \bar D$. \emph{Proof.} For each $k$, the pair $(\pi(k), \pi(k+1))$ is uniformly distributed over the $n(n-1)$ ordered pairs of distinct indices, so $\mathbb{E}_\pi\lVert p_{\pi(k+1)} - p_{\pi(k)}\rVert = \bar D$; averaging over $k$ gives the result. $\square$ After order permutation, mean saccade amplitude is therefore an unbiased, noisy measurement of spatial spread; it carries no sequential information. Section~\ref{sec:against} uses this result. The \emph{duration-coupling} transformation removes the assignment of durations to fixations. That assignment includes more than it may seem: any trend of duration along the trial and any serial dependence between successive durations are destroyed too, so a sensitivity to this transformation cannot say which of these carried the signal (the duration-trend simulation of Section~\ref{sec:sim} is a case in point). Because it preserves the trial's duration distribution, a descriptor that reads mean or total duration is invariant to it, and a task in which one class simply fixates longer would show zero sensitivity to it. The \emph{duration-marginal} transformation replaces the trial's durations by a draw without replacement from a donor pool, which destroys the trial's own distribution (and, with it, the coupling). The pool holds every fixation duration of the problem's \emph{other} participants, so no trial receives durations of its own participant. The pool is built once on the full dataset and is not rebuilt inside cross-validation or bootstrap draws, so held-out participants donate durations to trials in the training folds; the transformation is transductive in that limited sense. Rebuilding the pool from the training participants of each fold changed the classic descriptor's duration-marginal sensitivity by at most 0.007 AUC on the three problems where that sensitivity is interpreted, and by at most 0.064 on any problem, the largest changes falling on problems without signal whose intervals are wider than 0.3 (Table S24 in the Appendix). The \emph{rotation} transformation is a rigid motion about a centre: every intra-scanpath distance and every temporal relation is preserved, and so is every fixation's distance from the centre; it destroys orientation and side. Because half of the draws also reflect the scanpath, it destroys handedness as well: clockwise and counter-clockwise circulation, or left-to-right and right-to-left sweeps, are exchanged. We include the reflection so that a descriptor reading handedness cannot pass the transformation unchanged; the cost is that a rotation sensitivity alone does not separate orientation from handedness. The package therefore also offers the rotation without reflection. A class contrast between central and peripheral viewing survives either version entirely, which we demonstrate in Section~\ref{sec:sim}; the transformation is therefore a test of rotation/reflection sensitivity, not of stimulus location in general. For a descriptor whose spatial support is bounded (a triangulation of a stimulus's areas of interest), rotation about the frame centre can sweep fixations off the support and register as a spurious sensitivity, so we rotate such descriptors about the support's centre. That reduces the problem but does not remove it: a rectangle or an irregular set of anchors does not map onto itself under rotation, and jitter moves fixations near the boundary out of it as well. A fixation that falls outside the triangulation lies in no triangle and is dropped from the graph. We therefore report the share of fixations and dwell time each transformation leaves on the support, by class, and check the interpreted sensitivities with support-matched transformations (Section~\ref{sec:support}). The \emph{jitter} transformation perturbs each location by $\mathcal{N}(0, s^2 I)$; two originally coincident fixations end up on average $s\sqrt{\pi}$ apart (root-mean-square $2s$). Jitter preserves order and durations exactly and nothing else exactly: it is a gradual perturbation that changes pairwise distances, amplitudes, local density, orientation and region membership together, more at fine scales than at coarse ones. ``Spatial structure finer than $s$'' is therefore operational shorthand for what the transformation mostly disturbs, not a filter with a cut-off. Sweeping $s$ produces a curve of AUC against scale from which we report two summaries with explicit definitions (Section~\ref{sec:stat}).

\subsection{Sensitivity, and What It Is Not}\label{sec:stat}
For a transformation $\tau$ we draw $R$ independent replicates, recompute features and participant-held-out decision values on each, and set $\AUC_\tau = R^{-1}\sum_r \AUC(\tau_r)$. The \emph{sensitivity} of the descriptor to $\tau$ is $\Cs_\tau = \AUC_{\mathrm{obs}} - \AUC_\tau$. A negative sensitivity means that the transformed pipeline discriminates better than the observed one. It has two possible sources, and they must be told apart. The transformation can make information accessible that the observed features express poorly (the Proposition of Section~\ref{sec:transf} gives an exact case), or it can change the evaluation's bias, for example by moving an AUC that lies below its null distribution back towards that null. We separate them with three permutation distributions, built with the problem's own permutation scheme (Section~\ref{sec:inf}) on the same transformed trials: for the observed AUC, for the transformed AUC, and for the contrast $\Cs_\tau$ itself. A negative sensitivity is read as information made accessible when the transformed AUC exceeds its null and $\Cs_\tau$ lies below the null distribution of the contrast. It is read as a change in evaluation bias when the observed AUC lies below its null and the transformed AUC does not exceed its own. A significant observed AUC is therefore not required for a negative sensitivity to be informative.

The classifier is refit on the transformed trials, so $\Cs_\tau$ is the discrimination a \emph{pipeline} (descriptor plus learner plus evaluation) loses when the property is destroyed and it is allowed to adapt. It is not a measurement of what the originally fitted classifier relied on. A refitted learner can move to information that is redundant with the destroyed property, so a sensitivity of zero is consistent with the original model having leaned on that property heavily; only the pipeline's \emph{achievable} discrimination is shown to be unaffected. We use \emph{sensitivity} throughout in that pipeline sense.

Refitting follows established practice. \citet{hooker2019benchmark} retrain on the modified data because re-scoring the original model confounds the information removed with the artefact of moving off the training distribution, and \citet{hooker2021unrestricted} show that permuting a feature that depends on others forces that extrapolation. A perturbation profile applies the same remedy to whole trials, with one advantage: each transformation preserves stated invariants, so a transformed trial is still a scanpath, with the same number of fixations and, under three of the five transformations, the same set of locations. What the refit costs is the pipeline reading above.

Three caveats govern interpretation. First, $\Cs_\tau$ is a property of the descriptor, learner, evaluation, task, and transformation jointly; it is not a share of information in the behaviour, and $\sum_\tau \Cs_\tau$ need not equal the distance of $\AUC_{\mathrm{obs}}$ from its permutation null. Second, zero sensitivity is ambiguous between invariance of the descriptor, redundancy of the property with others the descriptor also reads (which refitting is free to exploit), insufficient precision, and information the transformation leaves intact. Third, the transformations are applied separately, not sequentially, so the profile does not partition the signal; interactions and redundancy are visible only by comparing descriptors that read different properties, which is how we use the baselines below.

For the jitter sweep, two summaries are used in the simulations of Section~\ref{sec:sim}, where the planted scale is known; for the real data we report the sweep itself. The \emph{first-exclusion scale} is the first tested grid point at which the pointwise 95\% interval for $\Cs$ excludes zero. It is a pointwise, unadjusted quantity: the intervals along the sweep are neither simultaneous nor corrected for the number of grid points, so it is not an estimate of where sensitivity begins, and it moves with the grid, the sample size and the Monte-Carlo settings. The \emph{half-loss bracket} is the pair of adjacent tested scales between which the point estimate $\Cs$ first exceeds half of $\AUC_{\mathrm{obs}} - \AUC_0$, where $\AUC_0$ is the mean of the observed AUC's permutation distribution (0.5 is not used as the reference, because the null is not centred there); it uses no interval. Neither locates a spatial scale of the behaviour. The primary result of a sweep is the curve of sensitivities with their intervals (Figure~\ref{fig:profiles}; Tables S1--S13 in the Appendix); the two summaries are reading aids for it.

\subsection{Inference}\label{sec:inf}
Trials are clustered within participants, and cross-validated AUCs have no unbiased variance estimator because folds share training data \citep{varoquaux2018cross}. Resampling participants over the \emph{fixed} out-of-fold scores (the fixed-score bootstrap) captures trial-sampling variance but not the variance of refitting the classifier, so its intervals are too narrow wherever refitting varies. We therefore use a \emph{refit bootstrap}: participants are resampled with replacement (duplicates keep their identity, so folds remain participant-pure), the classifier is refit by participant-held-out cross-validation on the resampled observed features and on one randomly chosen transformed replicate, and the two AUCs $a_{\mathrm{obs}}$ and $a_{\tau}$ are recorded over $B$ draws.

The estimator averages the transformed AUC over $R$ replicates, whereas each bootstrap draw uses one, so the bootstrap variance of $a_{\mathrm{obs}} - a_\tau$ overstates the estimator's variance. Write $a_\tau^{(b)} = g(\text{sample}_b) + e_b$, where $g$ is the participant-sampling component and $e_b$ the Monte-Carlo component of drawing one replicate, assumed independent of the participant resample with variance $v_e$ not depending on the resample. Then $\mathrm{Var}_b(a_\tau) = v_g + v_e$, whereas the estimator's transformed term has variance $v_g + v_e/R$, giving
\begin{equation}
\widehat{\mathrm{Var}}(\Cs_\tau) = \mathrm{Var}_b(a_{\mathrm{obs}}) + \big[\mathrm{Var}_b(a_{\tau}) - \hat v_e\big] + \hat v_e/R - 2\,\mathrm{Cov}_b(a_{\mathrm{obs}}, a_{\tau}),
\label{eq:var}
\end{equation}
where $\hat v_e$ is the sample variance of the $R$ replicate AUCs on the full data, capped at $\mathrm{Var}_b(a_\tau)$, and a normal-approximation 95\% interval is placed around $\Cs_\tau$. Two cases are handled separately. A cell is reported as ``0 (exact)'' with no interval when the transformation leaves the \emph{feature matrix} unchanged: when $f(\tau_r(\text{trial})) = f(\text{trial})$ to within $10^{-9}$ for every trial in every one of the $R$ replicates. The sensitivity is then identically zero under every resample and its sampling variance is zero. Every ``0 (exact)'' cell below is an invariance that holds analytically for the descriptor (recurrence reads only pairwise distances; the dispersion and count baselines read no order), verified by feature equality on the replicates drawn. A \emph{degenerate} cell, in which every bootstrap draw gives $a_{\mathrm{obs}} = a_\tau$ without the features being equal (which occurs at ceiling), is marked separately. If the corrected variance is not positive for a non-invariant cell, the uncorrected paired bootstrap variance $\mathrm{Var}_b(a_{\mathrm{obs}} - a_\tau)$ is used instead, which under the assumptions of Equation~\ref{eq:var} has expectation larger by $v_e(1 - 1/R)$; a larger variance estimate is not by itself a coverage guarantee, since it addresses neither bias nor the normal approximation, and the cell is flagged. Writing Equation~\ref{eq:var} under the constant-$v_e$ assumption does not verify it. Section~\ref{sec:sim} checks the coverage of the resulting interval by simulation and compares it with a \emph{direct} refit bootstrap that averages all $R$ transformed replicates inside every draw, which needs no correction but costs $R$ times as much.

The interval is built around $\Cs_\tau$ for the dataset in hand; what a simulation can check coverage against is
\begin{equation}
\theta_\tau = \mathbb{E}_D\big[\AUC_{\mathrm{obs}}(D) - \mathbb{E}_r \AUC_{\tau,r}(D)\big],
\label{eq:estimand}
\end{equation}
the expectation, over datasets $D$ from a generative design, of the finite-sample participant-held-out contrast. This is a \emph{procedural} estimand: it is defined by the design together with the participant count, the descriptor, the learner and the participant-held-out evaluation, so a study with more participants defines a different $\theta_\tau$ and the value estimated here transfers to no other sample size. Section~\ref{sec:sim} estimates $\theta_\tau$ on datasets whose seeds are disjoint from those whose intervals are assessed. Averaging replicate AUCs, rather than averaging decision values across replicates before computing an AUC, also matters: the latter ensembles the replicates into a different classifier, which can change $\AUC_\tau$ and, when it raises it, biases $\Cs_\tau$ downward.

All intervals below are unadjusted 95\% intervals and are exploratory: each profile comprises eight columns (five transformations, jitter at four scales), and the families across descriptors and problems are large. The estimates and their intervals are the primary result. The markers in the tables (an interval excluding zero; an interval inside the margin) are reading aids that dichotomize them, and we do not treat an isolated exclusion as a confirmed finding or an interval that covers zero as evidence of absence. For the latter we use an explicit equivalence margin: a sensitivity is called \emph{negligible} only when its whole interval lies within $\pm 0.05$ AUC; otherwise it is \emph{inconclusive}. An interval can both exclude zero and lie within the margin, in which case the effect is \emph{nonzero but negligible}. The margin is a judgement of practical relevance. We set it at $0.05$ AUC as the tolerance we would apply in this application (a shift of less than five hundredths of an AUC point is not, in our judgement, a reason to describe a descriptor as reading a property), and a reader with a different tolerance should reclassify accordingly. It is not justified by ranking preservation. The behaviour of the call near the margin was not re-examined by simulation for this pipeline, so we use it only as a reading aid. The margin is applied only to genuine intervals, never to exact invariances.

\paragraph{Permutation tests.} Each observed AUC is compared with a permutation distribution \citep{ojala2010permutation} of 1000 permutations ($p = (1 + \#\{\text{permuted AUC} \geq \text{observed}\})/1001$); its mean and SD and the lower-tail probability are stored with it. Labels are permuted at the level at which they were assigned and within blocks that keep the design's structure: whole participants for a label that is constant within participants, passages within participants for SB-SAT comprehension (the pages of a passage share its outcome), the two halves within each recording for the half problems, answers within each program for pooled correctness (so that the association of correctness with the program is preserved under the null), and trials within participants otherwise (Table~\ref{tab:perm}). The same schemes give the nulls of transformed AUCs and of contrasts.

\paragraph{Which sensitivities are interpreted.} A profile of eight columns for each of ten descriptor variants on thirteen problems produces several hundred intervals, and reading off those that exclude zero would capitalize on chance. We therefore fix a rule. A cell is \emph{interpreted} only if the observed AUC exceeds its permutation distribution ($p \leq 0.05$), the refit-bootstrap interval of the observed AUC lies above 0.5, and the sensitivity interval excludes zero. An interpreted cell is \emph{robust} if its sign and the exclusion of zero hold as well with the pooled AUC, in every one of 20 fold partitions (for the sign of the point estimate), in three longer runs ($R = 20$, $B = 300$) and, for a descriptor with bounded support, on a version of the descriptor that loses no fixation under any transformation. The result under a second learner is reported alongside. A negative sensitivity on a problem whose observed AUC does not exceed its null can still be informative by the three-null argument of Section~\ref{sec:stat}, but it is outside the rule and we report such cases without building on them.

\begin{table}[!htbp]
\centering\small
\caption{Permutation schemes. For each problem: the unit at which the label was assigned, the unit whose labels are permuted, the block within which they are permuted, and what every permutation preserves.}
\label{tab:perm}
\resizebox{\linewidth}{!}{\input{tab_perm}}
\end{table}

\subsection{Invariance Checks}
Some (descriptor, transformation) pairs must return $\Cs = 0$ exactly: order for any order-free descriptor; duration coupling and duration marginal for any descriptor that does not read durations; rotation for any descriptor that depends only on pairwise distances or on the distance matrix. These serve as checks that the machinery is correct, not as evidence about the data. They held in every analysis below.

%% file: tab_perm.tex
\begin{tabular}{>{\raggedright\arraybackslash}p{3.3cm}>{\raggedright\arraybackslash}p{2.6cm}>{\raggedright\arraybackslash}p{2.6cm}>{\raggedright\arraybackslash}p{2.4cm}>{\raggedright\arraybackslash}p{4.1cm}}
\toprule
Problems & Label assigned to & Unit permuted & Block & Preserved under the null \\
\midrule
\texttt{gb\_tex\_vd1}, \texttt{gb\_vd1\_blg}, \texttt{gb\_vd1\_vd2}, \texttt{gb\_tex\_s1s2}, \texttt{em\_program} & trial (task, stimulus or session) & trial & participant & each participant's label counts; participant clustering \\
\texttt{gb\_vd1\_halves}, \texttt{em\_halves} & half of one recording & half & recording (participant $\times$ session or program) & one first and one second half per recording \\
\texttt{sb\_native}, \texttt{em\_expertise}, \texttt{em\_correctrect}, \texttt{em\_correctveh} & participant & participant (all trials together) & all participants & the number of participants per class; all trials of a participant share one label \\
\texttt{sb\_compr} & passage (its pages share one outcome) & passage & participant & each participant's passage-label counts; pages of a passage share one label \\
\texttt{em\_correct} & answer to one program & trial & program & the number of correct and incorrect answers within each program \\
\bottomrule
\end{tabular}

%% file: sec_sim.tex
\section{Validation by Simulation}\label{sec:sim}
\subsection{Design}
The simulations use the pipeline of the real-data analyses: a linear SVM on standardized features and ten-fold cross-validation over participants, with the participants dealt to class-balanced folds in a fixed, seeded order, and the within-fold AUC. The recovery runs, whose labels vary within participants, used the pooled AUC. Each synthetic trial has 40 fixations in a $600\times500$ frame, and every participant carries a random offset. Scenarios plant a signal in one property: \emph{order} (same location multiset; one class visits its locations in a noisy left-to-right sweep), \emph{duration coupling} (same locations and order; durations increase along the trial), \emph{duration marginal} (durations shifted by 120 ms), \emph{side} (same intrinsic geometry, opposite corners), \emph{eccentricity} (central versus a ring at the same mean position), \emph{scale} (both classes revisit four clusters, within SD $\sigma$ or $3\sigma$, for $\sigma = 20$ and 60 px), and \emph{none}. Recovery was checked on 20 datasets of 20 participants $\times$ 8 trials ($R = 5$, $B = 60$), each scenario with a descriptor that can read the planted property and, for most, one that cannot.

Calibration was checked at sample sizes at or below those of the analysed problems: 60 participants $\times$ 4 trials and 112 $\times$ 2 with the label varying within participants, and 95 $\times$ 20 and 65 $\times$ 2 with the label constant within participants (the sizes of the EMIP and SB-SAT problems). A further campaign at 322 participants $\times$ 4 trials, the size of the GazeBase problems, covers the no-signal and signal-elsewhere designs. For the between-participant sizes each participant is assigned one class and keeps only trials of that class, so the class-conditional law of a trial is unchanged. Three designs were run: \emph{none}; \emph{strong signal elsewhere} (a spread contrast, $\sigma$ vs.\ $1.35\sigma$, with cluster centres uniform in a disc so that the design is rotation-symmetric; AUC about 0.8); and the same without the participant offset. The order target is zero by exchangeability in every design, and the rotation target is zero in the design without the offset. Every other target $\theta_\tau$ (Equation~\ref{eq:estimand}) was estimated on 60 further datasets per cell with disjoint seeds (point estimates, $R = 20$); we report the Monte Carlo standard error of each estimate. The campaign was run at the Monte Carlo settings of every reported profile, $R = 6$ and $B = 60$, with 40 datasets per design and size (Table~\ref{tab:cal}).

Three further checks address the assumptions of Equation~\ref{eq:var} and the range of designs. (i) On six cells of the calibration designs, the corrected interval was compared with the \emph{direct} refit bootstrap on the same 200 datasets per cell. (ii) New designs added a weak spread contrast (classes within $\sigma$ vs.\ $1.35\sigma$ around non-symmetric centres; observed AUC 0.64--0.88 depending on size), a weak order contrast (the sweep's sort noise raised from 40 to 150 px; AUC 0.60--0.63), and a design with a planted \emph{negative} sensitivity: one class sweeps a $600\times500$ field in noisy left-to-right order and the other visits a $345\times345$ field at random, chosen so that mean saccade amplitudes match in the observed order but, by the Proposition of Section~\ref{sec:transf}, not after permutation. (iii) The no-signal and weak-spread designs were rerun with unequal trial counts (one to four trials per class and participant), with class imbalance (one positive in four trials within participants; 16 of 65 participants positive between participants), and with strong participant heterogeneity (offset SD 100 instead of 25 px, and a participant-specific scaling of every scanpath by $e^{z}$, $z \sim \mathcal{N}(0, 0.3^2)$). Each new cell has 40 datasets and its own reference run (60 datasets, disjoint seeds).

The dataset is the unit of replication. A design and size contributes 40 datasets but several intervals per dataset (one per contrast and descriptor), so we report both counts; a contrast's rejections are counted over its 40 datasets, with the exact one-sided 95\% upper bound on the rate (7.2\% for 0 of 40), and pooled rates are means over datasets of the per-dataset rejection fraction.

\subsection{Results}
\emph{Recovery.} Every planted property was detected in 20 of 20 datasets by its reader: order ($+0.42$ for the directed graph, $+0.50$ for recurrence), duration coupling ($+0.50$), duration marginal ($+0.52$), side ($+0.52$) and spatial scale. The jitter curve tracked the planted scale: the half-loss bracket was $[40, 80]$ px when the classes returned within 20 vs.\ 60 px and $[160, 320]$ px when they returned within 60 vs.\ 180 px. The duration-trend descriptor also lost its whole signal under order permutation in the duration-coupling scenario ($+0.50$), which is correct rather than a failure: it reads durations as a function of sequence position, so destroying either property destroys its signal. The two blind spots behaved as stated: under a planted duration marginal, duration coupling returned exactly zero for a descriptor reading mean duration while the duration-marginal transformation recovered $+0.52$; under a planted eccentricity contrast, rotation returned exactly zero for a descriptor reading eccentricity although its AUC was 1.0. Across the control descriptors, which cannot read the planted property, 7 of 520 genuine intervals excluded zero (1.3\%), and every required invariance held exactly.

\emph{Calibration at the reported settings} (Table~\ref{tab:cal}). The 480 datasets of the calibration campaign gave 2560 genuine intervals. Of the 1280 whose target is zero by argument, 4 excluded zero (0.3\%); the pooled rate was at most 1.7\% $\pm$ 1.2\%, and no contrast exceeded 2 of 40 (upper bound 14.9\%). Against their targets, zero or estimated, 2553 of 2560 intervals covered (99.7\%), and the lowest coverage in any design and size was 98.3\%. The estimated targets were precise enough for this check: their Monte Carlo standard errors had a median of 0.002 and a maximum of 0.011, against mean interval widths of 0.02--0.46.

The intervals are therefore conservative, and by a wide margin. Across the 64 contrasts, the mean bootstrap standard error was 1.2 to 3.4 times the empirical SD of the estimate over datasets (median 1.8), and intervals built at nominal 80\% and 90\% covered in 96.2\% and 99.0\% (Table S17). A nominal 95\% interval from this procedure should be read as an interval with at least that coverage and with correspondingly less power: under no signal its mean width was 0.12 at 95 $\times$ 20, 0.35 at the two within-participant sizes and 0.46 at 65 $\times$ 2.

\emph{Calibration at 322 participants} (Table S26). At 322 $\times$ 4, 40 datasets per design gave 520 intervals. Two of the 200 with a zero target excluded zero (1.0\%), 516 of 520 covered their target (99.2\%; lowest for a single contrast 38 of 40), the standard-error ratio had a median of 1.6 (1.3--2.0), and the nominal 80\% interval covered in 95.6\%. The mean width under no signal was 0.14. The intervals are thus conservative at the sample size that carries most of the results below, somewhat less so than at the smaller sizes.

\emph{Direct bootstrap} (Table S18). On 2400 paired intervals (200 datasets for each of six cells), the direct bootstrap covered in 99.4\% and the corrected interval in 99.7\%; the corrected interval was 18--61\% wider, most of all where another property carried strong signal, and each excluded zero at a zero target in 6 of 1200 cases. The lowest coverage of any contrast was 98.5\% for either interval. The direct bootstrap was conservative too (standard-error ratios of 1.3--2.2, against 1.5--2.8 for the corrected interval), so most of the conservatism comes from resampling participants for a cross-validated statistic, and the correction of Equation~\ref{eq:var} adds to it; it did not make the interval anti-conservative in any tested cell.

\emph{Weak, negative, unequal, imbalanced and heterogeneous designs} (Table S19). Across 2000 intervals from 960 datasets, 1994 covered their target (99.7\%; lowest per contrast 92.5\%, 37 of 40, for the order contrast under no signal with imbalanced classes at 65 participants), and 3 of the 800 intervals with a zero target excluded zero. The planted negative sensitivity ($\theta = -0.50$ at both sizes, with observed AUCs of 0.51 and 0.52) was detected in all 80 datasets although the observed pipeline had no discrimination, which is the situation Section~\ref{sec:stat} describes. Weak effects show the cost of the conservatism: the weak order contrast ($\theta = 0.12$--$0.13$) excluded zero in 35\% of datasets at 60 $\times$ 4 and in 8\% at 65 $\times$ 2, and the weak spread contrast under strong heterogeneity ($\theta = 0.13$--$0.14$) in 60\% and 13\%, whereas sensitivities of 0.30 excluded zero in every dataset. Unequal trial counts and class imbalance widened the intervals, most at 65 participants with 16 positives (0.56 under no signal).

The choice of the within-fold AUC as the primary metric rests on the same simulations. Under no signal its mean was 0.50--0.51 at every size, and 0.47--0.51 under imbalance, unequal trial counts and heterogeneity. In an earlier run of the campaign with the pooled AUC and unbalanced folds, the mean observed AUC under no signal was 0.49--0.50 at the two within-participant sizes but 0.43--0.44 at the two between-participant sizes, and 0.39 with imbalanced classes: a cross-validated pooled AUC under a between-participant label sits below 0.5 when there is nothing to find. Every observed AUC below is nevertheless read against its permutation distribution, and every profile is repeated with the pooled AUC as a sensitivity analysis.

\begin{table}[!htbp]
\centering\small
\caption{Calibration of the refit-bootstrap interval at the Monte Carlo settings of the reported profiles ($R = 6$, $B = 60$; within-fold AUC, class-balanced folds). Per design and size (40 datasets each): intervals with target zero by argument, those excluding zero, the pooled rejection rate ($\pm$ SE over datasets), coverage of the target (zero or estimated) over all genuine intervals, and the mean interval width. W: label within participants; B: between participants.}
\label{tab:cal}
\input{tab_cal}
\end{table}

%% file: tab_cal.tex
\begin{tabular}{llrrrrr}
\toprule
Design & Size & Zero-target int. & Excl.\ 0 & Rate (\%) & Coverage & Width \\
\midrule
no signal & 60 $\times$ 4 (W) & 120 & 0 & 0.0 $\pm$ 0.0 & 1.000 & 0.351 \\
no signal & 112 $\times$ 2 (W) & 120 & 0 & 0.0 $\pm$ 0.0 & 1.000 & 0.354 \\
no signal & 95 $\times$ 20 (B) & 120 & 2 & 1.7 $\pm$ 1.2 & 0.988 & 0.115 \\
no signal & 65 $\times$ 2 (B) & 120 & 0 & 0.0 $\pm$ 0.0 & 0.997 & 0.457 \\
signal elsewhere & 60 $\times$ 4 (W) & 80 & 0 & 0.0 $\pm$ 0.0 & 1.000 & 0.167 \\
signal elsewhere & 112 $\times$ 2 (W) & 80 & 0 & 0.0 $\pm$ 0.0 & 1.000 & 0.169 \\
signal elsewhere & 95 $\times$ 20 (B) & 80 & 0 & 0.0 $\pm$ 0.0 & 1.000 & 0.037 \\
signal elsewhere & 65 $\times$ 2 (B) & 80 & 0 & 0.0 $\pm$ 0.0 & 1.000 & 0.252 \\
signal elsewhere, no offset & 60 $\times$ 4 (W) & 120 & 0 & 0.0 $\pm$ 0.0 & 1.000 & 0.111 \\
signal elsewhere, no offset & 112 $\times$ 2 (W) & 120 & 0 & 0.0 $\pm$ 0.0 & 1.000 & 0.114 \\
signal elsewhere, no offset & 95 $\times$ 20 (B) & 120 & 2 & 1.7 $\pm$ 1.2 & 0.983 & 0.023 \\
signal elsewhere, no offset & 65 $\times$ 2 (B) & 120 & 0 & 0.0 $\pm$ 0.0 & 1.000 & 0.188 \\
\bottomrule
\end{tabular}

%% file: tab_problems.tex
\begin{tabular}{lllrr}
\toprule
Key & Dataset & Contrast & Trials & Part. \\
\midrule
\texttt{gb\_tex\_vd1} & GazeBase & reading vs. video viewing (within) & 1288 & 322 \\
\texttt{gb\_vd1\_blg} & GazeBase & video viewing vs. ball-catching game (within) & 1287 & 322 \\
\texttt{gb\_vd1\_vd2} & GazeBase & video segment 1 vs. 2 (within) & 1288 & 322 \\
\texttt{gb\_vd1\_halves} & GazeBase & video 1, first vs. second half (within) & 1288 & 322 \\
\texttt{gb\_tex\_s1s2} & GazeBase & reading, session 1 vs. 2 (within) & 644 & 322 \\
\texttt{sb\_native} & SB-SAT & native vs. non-native English readers, reading pages (between) & 2046 & 95 \\
\texttt{sb\_compr} & SB-SAT & high vs. low passage comprehension, reading pages (mixed) & 1363 & 94 \\
\texttt{em\_program} & EMIP & Rectangle vs. Vehicle program (within) & 224 & 112 \\
\texttt{em\_halves} & EMIP & first vs. second half of a program reading (within) & 448 & 112 \\
\texttt{em\_expertise} & EMIP & high vs. low/no self-rated programming expertise (between) & 115 & 65 \\
\texttt{em\_correct} & EMIP & correct vs. incorrect answer, both programs pooled (confounded) (mixed) & 271 & 159 \\
\texttt{em\_correctrect} & EMIP & correct vs. incorrect answer, Rectangle program only (between) & 133 & 133 \\
\texttt{em\_correctveh} & EMIP & correct vs. incorrect answer, Vehicle program only (between) & 138 & 138 \\
\bottomrule
\end{tabular}

%% file: sec_results.tex
\section{Results}\label{sec:results}
\begin{figure}[!htbp]
\centering
\includegraphics[width=\linewidth]{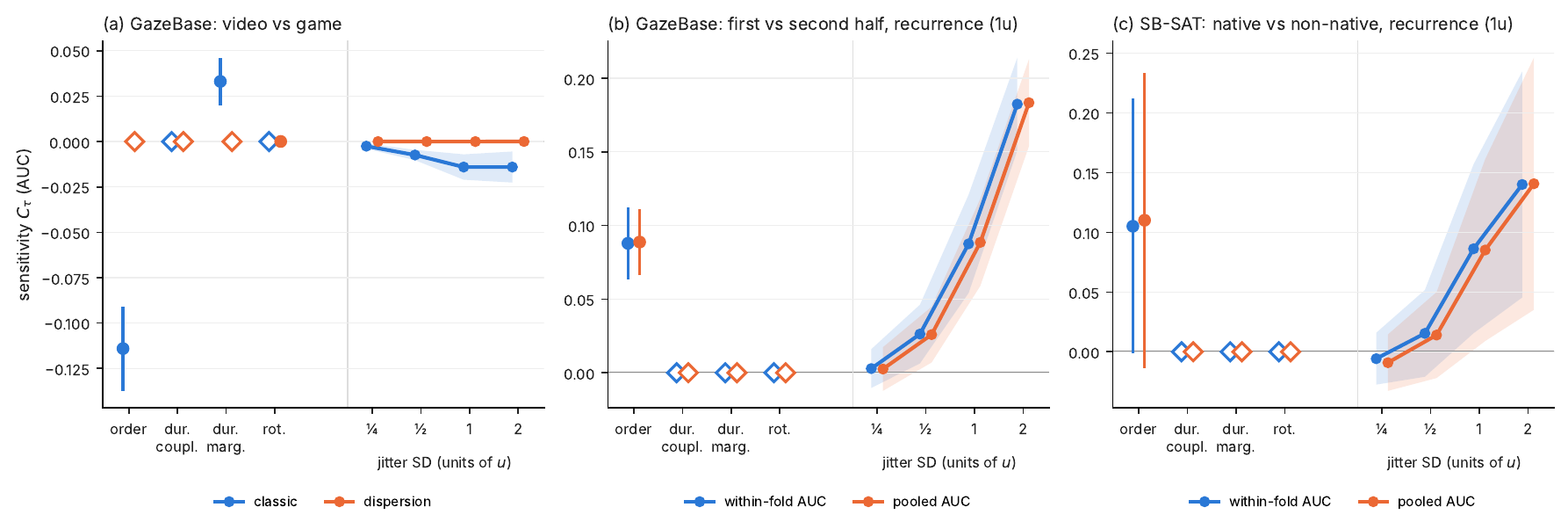}
\caption{Three perturbation profiles from the real data (within-fold AUC, class-balanced folds; $R = 6$, $B = 60$). Points: sensitivity $\Cs_\tau$ with its 95\% refit-bootstrap interval for the four transformations other than jitter; open diamonds: exact invariances (the transformed features equal the observed ones, so $\Cs_\tau = 0$ with no interval); curves and bands: the jitter sweep, with jitter SD in units of the reference scale $u$. (a) GazeBase video versus game: the classic descriptor gains discrimination when order is permuted, while the dispersion baseline is order-invariant by construction. (b) GazeBase first versus second half of a video: recurrence at $1u$ under the within-fold and the pooled AUC; the two profiles nearly coincide. (c) SB-SAT native versus non-native readers, recurrence at $1u$: the jitter sensitivity at $1u$ and $2u$ excludes zero under both metrics, whereas the order sensitivity does not.}
\label{fig:profiles}
\end{figure}

Each descriptor was profiled on each problem with $R = 6$ transformed replicates and $B = 60$ refit-bootstrap draws (seed 42). The primary analysis uses the within-fold AUC with class-balanced participant folds; the same profiles with the pooled AUC are a sensitivity analysis. Table~\ref{tab:auc} gives the observed AUCs with their permutation $p$-values, and Tables S1--S13 in the Appendix give every profile with its intervals. Figure~\ref{fig:profiles} shows three of the profiles.

\subsection{Invariance Checks}
Every analytically required invariance held exactly: every transformation for count; order for dispersion and for the undirected graph descriptors; both duration transformations for every descriptor that does not read durations, and duration coupling for the classic descriptor, which reads only their mean; rotation for recurrence and for the classic descriptor. In each case the transformed feature matrix equalled the observed one in every replicate.

\begin{table}[!htbp]
\centering\small
\caption{Observed participant-held-out AUC per problem and descriptor (within-fold AUC; class-balanced folds; linear SVM; ten-fold cross-validation over participants), with the permutation $p$-value in brackets (1000 permutations under the schemes of Table~\ref{tab:perm}; smallest attainable value 0.001). On \texttt{em\_correct} and \texttt{sb\_compr} the permutation scheme preserves part of the stimulus structure, so the permutation mean can lie well above 0.5 (Table S15) and an AUC must be read against its $p$-value. Pooled AUCs are in Table S14. A dash: descriptor not applicable.}
\label{tab:auc}
\resizebox{\linewidth}{!}{\input{tab_auc}}
\end{table}

\subsection{Which Cells Are Interpreted}\label{sec:ledger}
The thirteen problems gave 116 profiles and 565 genuine intervals. To keep interpretation from following the intervals that happen to exclude zero, a rule was fixed before the text below was written. A cell (problem, descriptor, transformation) is \emph{interpreted} only if the observed AUC exceeds its permutation distribution ($p \leq 0.05$), the refit-bootstrap interval of the observed AUC lies above 0.5, and the sensitivity interval excludes zero. An interpreted cell is called \emph{robust} if its sign and the exclusion of zero also hold (i) with the pooled AUC, (ii) in each of 20 different fold partitions (sign of the point estimate), (iii) in each of three longer runs with $R = 20$ and $B = 300$ (seeds 42, 43, 44) and, (iv) for the bounded AOI graph, on a padded graph that loses no fixation under any transformation (Section~\ref{sec:graph}). Whether the cell also holds with gradient-boosted trees is reported next to it as a separate question, because it concerns the learner rather than the evidence for the cell.

Of the 565 intervals, 95 excluded zero, 89 cells were interpreted, 80 were robust and 45 of those also excluded zero with the same sign under the trees (Table~\ref{tab:ledger}; every cell with all its checks is in Table S16). The six cells that excluded zero but were not interpreted had observed AUCs that did not exceed their null; two of them are discussed in Sections~\ref{sec:against} and~\ref{sec:confound}. Seventy-four of the 80 robust cells are on GazeBase, whose 322 participants give intervals less than half as wide as those on SB-SAT and about a fifth as wide as those on EMIP (mean widths 0.05, 0.12 and 0.26). EMIP and SB-SAT each contribute three robust cells, and the text below does not interpret any EMIP or SB-SAT cell outside those six. Thirty-six of the 80 robust intervals lie wholly inside the $\pm 0.05$ margin: with 322 participants, effects of a few thousandths of an AUC are resolved, and we count them as nonzero but negligible. Table~\ref{tab:robust} lists the robust cells outside GazeBase and the largest ones within it.

\begin{table}[!htbp]
\centering\small
\caption{Evidence ledger by problem. Intervals: genuine sensitivity intervals over all descriptors (exact invariances excluded). Excluding zero: in the primary analysis. Interpreted, robust: as defined in Section~\ref{sec:ledger}. Both learners: robust cells whose interval also excludes zero with the same sign under gradient-boosted trees.}
\label{tab:ledger}
\input{tab_ledger}
\end{table}

\begin{table}[!htbp]
\centering\small
\caption{Robust cells: all those on SB-SAT and EMIP, and those on GazeBase that hold under both learners with a sensitivity of at least 0.10 in absolute value. AUC: observed within-fold AUC. Normalized: the sensitivity divided by the observed AUC minus the mean of its permutation distribution. Trees: the interval also excludes zero with the same sign under gradient-boosted trees.}
\label{tab:robust}
\resizebox{\linewidth}{!}{\input{tab_robust}}
\end{table}

\subsection{Within-Fold and Pooled AUC}\label{sec:fold}
With the within-fold AUC and class-balanced folds, the mean of the permutation distribution was 0.49--0.51 on every problem whose labels are exchangeable across the units permuted (Table S15). With the pooled AUC it was lower wherever the label is constant within participants: 0.48--0.49 on native versus non-native readers, 0.47--0.49 on expertise and on the two within-program correctness problems. The bias is smaller than in the simulations because class-balanced folds, used for both metrics here, remove most of it; with unbalanced folds the pooled permutation mean on native versus non-native readers was 0.41. Two problems have permutation means above 0.5 under both metrics, and for a reason that is not evaluation bias. On pooled correctness, labels are permuted within programs, so program identity survives every permutation, and a descriptor that reads the program keeps its discrimination under the null (mean 0.71 for dispersion). On passage comprehension, labels are permuted between the passages of a participant, and passage-level outcome frequencies survive (means up to 0.62).

Which observed AUCs differed from their nulls did not depend on the metric: 61 of 116 problem--descriptor pairs had $p < 0.05$ under each, and the two pairs that changed status had $p$-values between 0.01 and 0.11 under both. The sensitivities agreed closely. Over all 565 intervals the two point estimates differed by a median of 0.002 and at most 0.051; 90 intervals excluded zero under both metrics, 5 only with the within-fold AUC and 13 only with the pooled AUC, whose intervals were about 10\% narrower. Normalized by the discrimination available above the permutation mean, the robust sensitivities differed between the metrics by a median of 0.002 and at most 0.05. Only three interpreted cells failed the pooled check, each with an interval that ended within 0.005 of zero.

\subsection{When a Transformation Changes What a Feature Measures}\label{sec:against}
On video versus game in GazeBase, permuting fixation order \emph{raised} the AUC of the classic descriptor from 0.886 to 1.000 (sensitivity $-0.114$ $[-0.137, -0.091]$; under the trees $-0.118$ $[-0.148, -0.088]$). The Proposition of Section~\ref{sec:transf} gives the mechanism. Mean saccade amplitude is the mean distance between \emph{consecutive} fixations; after permutation its expectation is the trial's mean pairwise distance, a measure of spatial spread. Where the classes differ more in spread than in the lengths of their actual saccades, the permuted descriptor discriminates better. The dispersion baseline, which contains the mean pairwise distance, reached 1.000 on this problem. The three permutation distributions of Section~\ref{sec:stat} agree (Table S25): the observed AUC, the transformed AUC and the contrast each differed from their nulls ($p = 0.001$), so the transformation made information accessible and did not merely shift an evaluation bias. The same sign appeared, robustly under the SVM, on reading versus video ($-0.022$ $[-0.034, -0.010]$; also under the trees, although this small contrast did not differ from its own permutation distribution, $p = 0.14$) and on the two halves of a video (0.564 to 0.654; $-0.090$ $[-0.110, -0.069]$), where the dispersion baseline reached 0.924; under the trees the halves interval reached zero ($-0.055$ $[-0.114, +0.003]$).

The EMIP program contrast showed the largest effect of this kind (0.539 to 0.788; $-0.249$ $[-0.341, -0.157]$), but its observed AUC did not exceed its permutation distribution ($p = 0.10$), so by the rule above it is reported and not interpreted. The opposite case occurred on EMIP's first versus second half of a program reading, where the classic descriptor's AUC was 0.776 and permuting order cost $+0.172$ $[+0.112, +0.231]$, robustly and under both learners: there the actual saccades, not the spread, carry the difference between early and late reading.

The classic descriptor also showed small negative jitter sensitivities on the two ceiling problems ($-0.003$ to $-0.014$). They are robust under the SVM, negligible by the margin, and absent under the trees, and we attach no interpretation to them beyond the fact that 322 participants resolve them.

\subsection{A Stimulus Confound}\label{sec:confound}
Pooled over both programs, ``correct versus incorrect answer'' in EMIP reached an AUC of 0.716 with the dispersion baseline and 0.57--0.66 with the graph descriptors. Among the retained trials, correctness and program are strongly associated: 92 of 133 Rectangle answers were correct and 106 of 138 Vehicle answers incorrect. Three checks show that the pooled discrimination is discrimination of the program. First, when labels are permuted within programs, the dispersion baseline's permutation mean is 0.710, and its observed AUC does not exceed it ($p = 0.38$). Second, an AUC computed only over pairs of a correct and an incorrect trial of the \emph{same} program and the same test fold was 0.490 $[0.398, 0.581]$ for dispersion and between 0.49 and 0.58 for the other nine descriptors; the largest, for the lattice graph (0.577 $[0.484, 0.685]$; $p = 0.042$ unadjusted), has an interval that includes 0.5 and does not survive adjustment for ten descriptors (Table S20). The lattice graph is also the only descriptor whose all-pairs AUC exceeded the within-program permutation distribution (0.664 against a permutation mean of 0.566; $p = 0.003$), so a weak within-program signal for it cannot be excluded; we treat it as a lead. Third, within each program separately, no descriptor exceeded its permutation distribution after such an adjustment.

That the program is easy to read from the spread is shown by the program contrast itself (dispersion 0.957) and by a control in which every trial is replaced by the same number of fixations drawn uniformly over the bounding box of its program's code, so that nothing of the participant's gaze remains: dispersion then reached 1.000 and the two graph descriptors 0.86--0.88, above their values on real gaze (Table S20). The program contrast in EMIP is therefore a contrast between two layouts, and no descriptor's AUC on it says anything about how the programs were read. The one robust cell on this problem, the rotation sensitivity of the dispersion baseline ($+0.139$ $[+0.078, +0.199]$), is consistent with that reading: the two code blocks differ in aspect ratio, and rotation mixes the axes along which the baseline measures extent.

Stimulus confounding is thus a sufficient explanation of the pooled correctness signal. It is not proof that comprehension leaves no trace in gaze: the within-program problems have 133 and 138 participants with one trial each, and a weak separation would not be resolved at that size.

\subsection{Reading in a First and a Second Language}\label{sec:native}
Native and non-native English readers in SB-SAT were discriminated at an AUC of 0.689 by the classic descriptor, 0.669 by recurrence at $1u$, 0.654 by the duration-trend descriptor, 0.624 by count and 0.614 by recurrence at $0.5u$ and by the directed graph (all $p \leq 0.008$), and at 0.587 by the lattice graph ($p = 0.023$); dispersion did not differ from its permutation distribution (0.545; $p = 0.17$).

Three cells are robust, all of them jitter sensitivities of recurrence: at $1u$ radius, jitter of two lines cost $+0.140$ $[+0.045, +0.235]$ and jitter of one line $+0.086$ $[+0.016, +0.157]$; at $0.5u$ radius, jitter of two lines cost $+0.088$ $[+0.012, +0.164]$. Relative to the discrimination above the permutation mean these are losses of about four fifths, one half and four fifths. Only the first also held under the trees at the screening settings; at the long settings the second did as well (Section~\ref{sec:learners}). The order sensitivity of recurrence, which an earlier pooled analysis with unbalanced folds had suggested, did not exclude zero under either metric here, and none of the classic descriptor's intervals did.

Destroying spatial structure at the scale of one to two lines thus reduced recurrence's discrimination of first- and second-language reading, robustly for a linear learner and at two lines for both learners. That recurrence reflects line-scale rereading or regressions is a plausible hypothesis, not a result: jitter changes several spatial properties at once, and recurrence statistics do not identify regressions uniquely. Because the label is constant within participants, this is also a contrast between two groups of readers that cannot separate language background from any other difference between them.

\subsection{Tasks, Time on Task and Spatial Scale in GazeBase}\label{sec:gb}
With all 322 participants of Round~1, the four GazeBase contrasts that carry signal yield most of the robust cells, and three patterns among them are large enough to matter.

\emph{Durations.} The duration-trend descriptor (the correlation of duration with sequence position, and the mean duration) lost all of its discrimination when the duration distribution was replaced: $+0.473$ $[+0.442, +0.503]$ on reading versus video, $+0.252$ $[+0.199, +0.304]$ on video versus game, and $+0.286$ $[+0.227, +0.345]$ on the two video segments, each a normalized loss of about 1.0. On the two video segments it also lost about half under order permutation and under duration coupling ($+0.157$ and $+0.158$), which is what a descriptor that reads durations as a function of position should do, and it shows that the two segments differ in how fixation durations evolve over the minute as well as in their level. The classic descriptor's discrimination of the segments rested on duration alone ($+0.124$ $[+0.099, +0.150]$ under the duration-marginal transformation; normalized 0.88).

\emph{Coarse spatial structure.} On the two halves of a video and on the two video segments, every structured descriptor lost discrimination under jitter of $2u$ ($4^\circ$): on the halves, $+0.191$ for the directed graph, $+0.182$ and $+0.143$ for recurrence at $1u$ and $0.5u$, and $+0.130$ for dispersion; on the segments, $+0.108$ to $+0.118$ for recurrence and dispersion and $+0.069$ for the directed graph. These are normalized losses of 0.3 to 0.6, and all held under both learners. The losses at $1u$ were about half as large and those at $0.5u$ at most 0.033, so what separates early from late viewing of this trailer, as far as these descriptors read it, is spatial structure coarser than about $2^\circ$.

\emph{Extent along the screen axes.} The dispersion baseline reached 0.924 on the halves and lost $+0.249$ $[+0.224, +0.275]$ under rotation. The halves therefore differ in the extent of the gaze spread along the horizontal and vertical axes; this baseline is translation-invariant and cannot say whether they also differ in location.

Reading in session 1 versus session 2, which also changes the passage, was near chance for most descriptors (0.52--0.59); the two intervals that excluded zero there did not hold in the longer runs.

\subsection{Problems Without Signal}\label{sec:nosignal}
Passage comprehension in SB-SAT, correctness within the Vehicle program and expertise in EMIP showed no discrimination that survived their permutation tests after adjustment for ten descriptors. On comprehension the observed AUCs (0.41--0.59) lay in the lower half of their permutation distributions for every descriptor (upper-tail $p$ from 0.76 to 0.99). That null keeps each participant's number of low-scoring passages and so retains passage-level structure, with means up to 0.62; it tests only whether gaze on a passage is associated with the outcome on that passage. We have no explanation for the lower-tail position and do not interpret this problem. Expertise needs a separate statement, because its observed AUCs were not merely at chance but below it for nine of ten descriptors (0.26 for the AOI graph, 0.37 for the directed graph), and a below-chance cross-validated AUC invites a search for an explanation. We looked for one and report what we found (Table S21). The classifiers fitted on nine tenths of the participants reached training AUCs of only 0.55--0.57 on average, so there is little to learn; the test-fold AUCs were below 0.5 in six to nine of the ten folds; and the permutation distribution, which is centred at 0.49--0.50 for this problem, placed the AOI graph's value in its lower tail ($p_{\mathrm{lower}} = 0.007$ unadjusted) without a mechanism that we could identify. Systematically below-chance accuracy is known to arise in small samples when a classifier fitted on noise is tested on held-out cases whose class means are, by construction, shifted away from the training means \citep{jamalabadi2016classification}. We therefore report the expertise profiles in Table S10 and interpret none of them: 65 participants with one or two trials each cannot support a claim in either direction.

\subsection{The Graph Descriptors}\label{sec:graph}
The Fixation Adjacency Graph is undirected and order-free by construction, and its order sensitivity was exactly zero on every problem. On the stimulus-independent lattice it was competitive at ceiling (0.988 on reading versus video, 0.957 on video versus game) and within the range of the other structured descriptors elsewhere. Its robust cells are those of a coarse spatial descriptor: jitter of $2u$ cost it $+0.088$ on the GazeBase halves and $+0.049$ to $+0.055$ on reading versus video and on the two segments, and rotation $+0.029$ on video versus game.

On the code tokens of EMIP, the AOI graph reached 0.778 on the program contrast and 0.682 on the two halves of a program reading. Neither number supports the claim that anchoring the graph to the stimulus helps. On the program contrast the layout alone gives the AOI graph 0.879 (Section~\ref{sec:confound}). On the halves, two of its sensitivities excluded zero and held in the pooled, partition and long-run checks: rotation ($+0.098$ $[+0.031, +0.166]$) and jitter of two lines ($+0.113$ $[+0.014, +0.212]$).

\paragraph{Support retention.}\label{sec:support} Both of those transformations move fixations off the AOI triangulation, where the graph ignores them (Table S22). Observed trials kept 96\% (first halves) and 99\% (second halves) of their dwell on the support; rotation lowered this to 79\% and 89\%, and jitter of two lines to 83\% and 87\%. A descriptor made only of the two retention shares separated the halves better than the graph did (0.773) and lost $+0.123$ $[+0.054, +0.192]$ under jitter of two lines, because the amount of looking that leaves the code is itself different between halves. We therefore built a padded AOI graph, in which lattice anchors $2u$ apart surround the token anchors so that no transformation moves a fixation off the support (retention 1.000 in every cell). On the padded graph the observed AUC was 0.669, and neither sensitivity excluded zero: rotation $+0.047$ $[-0.017, +0.111]$ and jitter of two lines $+0.062$ $[-0.017, +0.141]$. About half of each default sensitivity was thus differential support loss, and what remains is not resolved at this sample size. Support-matched transformations, redrawn until a trial keeps its observed dwell share on the support to within 0.02, agree for rotation ($+0.032$ $[-0.059, +0.124]$; matched in 99\% of draws) and are uninformative at two lines of jitter, where a matching draw was found in only 31\% of cases (Table S23). Both cells fail the fourth check and are not counted as robust.

For the AOI graph, then, this analysis supports no positive claim. It discriminated the two programs by their layout, the halves of a reading partly by how much gaze left the code, and correctness within the Rectangle program at 0.658 ($p = 0.012$ unadjusted, with an interval for the AUC that includes 0.5), which is a lead for a confirmatory test and nothing more. Whether a graph on semantically meaningful anchors adds to simpler descriptors remains open; these data cannot answer it, because the two EMIP stimuli differ in layout and the within-stimulus problems are small.

\subsection{Longer Runs and Fold Partitions}\label{sec:conv}
The 89 interpreted cells were re-run with $R = 20$ and $B = 300$ under seeds 42, 43 and 44. The point estimates moved by a median of 0.001 and at most 0.029 from the main run and by at most 0.019 across seeds, and the mean interval width was 0.052 in the main run and 0.055 in the longer runs. Eighty-two cells excluded zero with the same sign in all three longer runs. The seven that did not had main-run intervals ending within 0.02 of zero. Across 20 fold partitions, every interpreted cell kept its sign in every partition, and the SD of the point estimate over partitions was 0.004--0.008 on EMIP and SB-SAT, an order of magnitude below the interval half-widths there. Fold assignment is therefore not a material source of variation in these results.

\subsection{A Second Learner}\label{sec:learners}
Because a profile describes a pipeline, the interpreted cells were repeated with gradient-boosted trees (XGBoost; 200 trees of depth 3, learning rate 0.05, row subsampling 0.8, fixed in advance rather than tuned; $R = 6$, $B = 60$). Of the 80 robust cells, 78 kept their sign under the trees and 45 had an interval that excluded zero. Keeping a sign is weaker than keeping statistical support, so we count only the 45 as holding under both learners. For the 21 cells of Table~\ref{tab:robust} the trees were also run at the long settings ($R = 20$, $B = 300$; Table S27): the point estimates moved by at most 0.020 and 20 verdicts were unchanged; the remaining cell, recurrence at $1u$ under one line of jitter on native versus non-native readers, excluded zero at the long settings ($+0.082$ $[+0.012, +0.151]$) and not at the screening ones ($+0.071$ $[-0.003, +0.145]$). The counts in Table~\ref{tab:ledger} use the screening settings throughout. The pattern of what was lost is informative. Of the 61 robust positive sensitivities, 43 held under the trees, including every one of at least 0.10. Of the 19 robust negative sensitivities, only two did, both of them the order sensitivity of the classic descriptor that the Proposition explains (video versus game, and reading versus video). Of the 17 robust negative cells that did not hold under the trees, 15 are small (at most 0.020 in absolute value), most of them on the two problems near ceiling, where a linear boundary gains a little on perturbed features and a more flexible learner has no room to reproduce the gain. Two are larger. The classic descriptor's order cell on the GazeBase halves ($-0.090$) has the mechanism of Section~\ref{sec:against}, and its interval under the trees reached zero. The other has no known mechanism: on the two halves of an EMIP program reading, recurrence at $0.5u$ (AUC 0.613) \emph{gained} 0.085 under jitter of one line ($-0.085$ $[-0.152, -0.018]$; normalized $-0.77$). Its contrast lay in the lower tail of its permutation distribution ($p = 0.027$; Table S25). It passed the pooled, partition and long-run checks and is therefore listed in Table~\ref{tab:robust}, but it failed under the trees, no neighbouring scale shows it, and we know of no mechanism for it. We report it as unexplained and build nothing on it. A sensitivity belongs to the descriptor and the learner together, and a negative one especially so.

%% file: tab_auc.tex
\begin{tabular}{lrrrrrrrrrr}
\toprule
Problem & Count & Dispersion & Classic & Dur.\ trend & RQA $0.5u$ & RQA $1u$ & Directed & FAG lattice & FAG AOI & FAG AOI pad. \\
\midrule
\texttt{gb\_tex\_vd1} & 0.969 (.001) & 1.000 (.001) & 0.969 (.001) & 0.957 (.001) & 0.999 (.001) & 1.000 (.001) & 0.977 (.001) & 0.988 (.001) & -- & -- \\
\texttt{gb\_vd1\_blg} & 0.690 (.001) & 1.000 (.001) & 0.886 (.001) & 0.752 (.001) & 0.999 (.001) & 1.000 (.001) & 1.000 (.001) & 0.957 (.001) & -- & -- \\
\texttt{gb\_vd1\_vd2} & 0.634 (.001) & 0.707 (.001) & 0.640 (.001) & 0.790 (.001) & 0.677 (.001) & 0.719 (.001) & 0.700 (.001) & 0.664 (.001) & -- & -- \\
\texttt{gb\_vd1\_halves} & 0.500 (1.000) & 0.924 (.001) & 0.564 (.001) & 0.590 (.001) & 0.743 (.001) & 0.823 (.001) & 0.805 (.001) & 0.741 (.001) & -- & -- \\
\texttt{gb\_tex\_s1s2} & 0.530 (.013) & 0.519 (.243) & 0.539 (.003) & 0.549 (.010) & 0.576 (.001) & 0.521 (.134) & 0.531 (.090) & 0.591 (.001) & -- & -- \\
\texttt{sb\_native} & 0.624 (.003) & 0.545 (.169) & 0.689 (.001) & 0.654 (.005) & 0.614 (.005) & 0.669 (.002) & 0.614 (.008) & 0.587 (.023) & -- & -- \\
\texttt{sb\_compr} & 0.554 (.763) & 0.413 (.975) & 0.587 (.990) & 0.575 (.929) & 0.497 (.914) & 0.529 (.919) & 0.533 (.897) & 0.506 (.810) & -- & -- \\
\texttt{em\_program} & 0.561 (.014) & 0.957 (.001) & 0.539 (.104) & 0.540 (.227) & 0.580 (.025) & 0.547 (.140) & 0.655 (.001) & 0.658 (.001) & 0.778 (.001) & 0.729 (.001) \\
\texttt{em\_halves} & 0.500 (1.000) & 0.616 (.001) & 0.776 (.001) & 0.582 (.009) & 0.613 (.001) & 0.680 (.001) & 0.634 (.001) & 0.599 (.001) & 0.682 (.001) & 0.669 (.001) \\
\texttt{em\_expertise} & 0.472 (.591) & 0.469 (.683) & 0.438 (.748) & 0.541 (.409) & 0.419 (.794) & 0.416 (.803) & 0.366 (.886) & 0.447 (.756) & 0.259 (.994) & 0.459 (.649) \\
\texttt{em\_correct} & 0.495 (.706) & 0.716 (.383) & 0.501 (.583) & 0.511 (.498) & 0.571 (.113) & 0.525 (.359) & 0.580 (.207) & 0.664 (.003) & 0.635 (.324) & 0.571 (.726) \\
\texttt{em\_correctrect} & 0.589 (.120) & 0.472 (.687) & 0.523 (.447) & 0.507 (.556) & 0.451 (.760) & 0.462 (.749) & 0.533 (.364) & 0.578 (.149) & 0.658 (.012) & 0.544 (.297) \\
\texttt{em\_correctveh} & 0.444 (.751) & 0.476 (.634) & 0.500 (.572) & 0.521 (.460) & 0.518 (.444) & 0.559 (.255) & 0.618 (.060) & 0.574 (.196) & 0.503 (.550) & 0.470 (.664) \\
\bottomrule
\end{tabular}

%% file: tab_ledger.tex
\begin{tabular}{lrrrrr}
\toprule
Problem & Intervals & Excluding zero & Interpreted & Robust & Robust, both learners \\
\midrule
\texttt{gb\_tex\_vd1} & 37 & 20 & 20 & 20 & 9 \\
\texttt{gb\_vd1\_blg} & 37 & 16 & 16 & 15 & 8 \\
\texttt{gb\_vd1\_vd2} & 37 & 21 & 21 & 19 & 13 \\
\texttt{gb\_vd1\_halves} & 37 & 20 & 20 & 20 & 12 \\
\texttt{gb\_tex\_s1s2} & 37 & 2 & 2 & 0 & 0 \\
\texttt{sb\_native} & 37 & 5 & 4 & 3 & 1 \\
\texttt{sb\_compr} & 37 & 1 & 0 & 0 & 0 \\
\texttt{em\_program} & 51 & 2 & 1 & 1 & 1 \\
\texttt{em\_halves} & 51 & 6 & 5 & 2 & 1 \\
\texttt{em\_expertise} & 51 & 0 & 0 & 0 & 0 \\
\texttt{em\_correct} & 51 & 1 & 0 & 0 & 0 \\
\texttt{em\_correctrect} & 51 & 1 & 0 & 0 & 0 \\
\texttt{em\_correctveh} & 51 & 0 & 0 & 0 & 0 \\
\midrule
Total & 565 & 95 & 89 & 80 & 45 \\
\bottomrule
\end{tabular}

%% file: tab_robust.tex
\begin{tabular}{lllrrrc}
\toprule
Problem & Descriptor & Transformation & AUC & Sensitivity [95\% interval] & Normalized & Trees \\
\midrule
\texttt{gb\_tex\_vd1} & Dur.\ trend & duration marginal & 0.957 & +0.473 [+0.442, +0.503] & $+1.03$ & $\checkmark$ \\
\texttt{gb\_vd1\_blg} & Classic & order & 0.886 & $-$0.114 [$-$0.137, $-$0.091] & $-0.30$ & $\checkmark$ \\
\texttt{gb\_vd1\_blg} & Dur.\ trend & duration marginal & 0.752 & +0.252 [+0.199, +0.304] & $+1.01$ & $\checkmark$ \\
\texttt{gb\_vd1\_vd2} & Classic & duration marginal & 0.640 & +0.124 [+0.099, +0.150] & $+0.88$ & $\checkmark$ \\
\texttt{gb\_vd1\_vd2} & Dispersion & jitter $2u$ & 0.707 & +0.116 [+0.085, +0.148] & $+0.56$ & $\checkmark$ \\
\texttt{gb\_vd1\_vd2} & Dur.\ trend & duration coupling & 0.790 & +0.158 [+0.131, +0.185] & $+0.54$ & $\checkmark$ \\
\texttt{gb\_vd1\_vd2} & Dur.\ trend & duration marginal & 0.790 & +0.286 [+0.227, +0.345] & $+0.98$ & $\checkmark$ \\
\texttt{gb\_vd1\_vd2} & Dur.\ trend & order & 0.790 & +0.157 [+0.127, +0.187] & $+0.54$ & $\checkmark$ \\
\texttt{gb\_vd1\_vd2} & RQA $0.5u$ & jitter $2u$ & 0.677 & +0.108 [+0.071, +0.145] & $+0.61$ & $\checkmark$ \\
\texttt{gb\_vd1\_vd2} & RQA $1u$ & jitter $2u$ & 0.719 & +0.118 [+0.080, +0.156] & $+0.54$ & $\checkmark$ \\
\texttt{gb\_vd1\_halves} & Directed & jitter $2u$ & 0.805 & +0.191 [+0.161, +0.220] & $+0.62$ & $\checkmark$ \\
\texttt{gb\_vd1\_halves} & Dispersion & jitter $2u$ & 0.924 & +0.130 [+0.111, +0.150] & $+0.31$ & $\checkmark$ \\
\texttt{gb\_vd1\_halves} & Dispersion & rotation & 0.924 & +0.249 [+0.224, +0.275] & $+0.59$ & $\checkmark$ \\
\texttt{gb\_vd1\_halves} & RQA $0.5u$ & jitter $2u$ & 0.743 & +0.143 [+0.097, +0.189] & $+0.59$ & $\checkmark$ \\
\texttt{gb\_vd1\_halves} & RQA $1u$ & jitter $2u$ & 0.823 & +0.182 [+0.151, +0.214] & $+0.56$ & $\checkmark$ \\
\texttt{sb\_native} & RQA $0.5u$ & jitter $2u$ & 0.614 & +0.088 [+0.012, +0.164] & $+0.79$ & -- \\
\texttt{sb\_native} & RQA $1u$ & jitter $2u$ & 0.669 & +0.140 [+0.045, +0.235] & $+0.85$ & $\checkmark$ \\
\texttt{sb\_native} & RQA $1u$ & jitter $1u$ & 0.669 & +0.086 [+0.016, +0.157] & $+0.52$ & -- \\
\texttt{em\_program} & Dispersion & rotation & 0.957 & +0.139 [+0.078, +0.199] & $+0.30$ & $\checkmark$ \\
\texttt{em\_halves} & Classic & order & 0.776 & +0.172 [+0.112, +0.231] & $+0.62$ & $\checkmark$ \\
\texttt{em\_halves} & RQA $0.5u$ & jitter $1u$ & 0.613 & $-$0.085 [$-$0.152, $-$0.018] & $-0.77$ & -- \\
\bottomrule
\end{tabular}

%% file: supp_profiles.tex
\begin{table}[!htbp]\centering\scriptsize
\caption{Profile of \texttt{gb\_tex\_vd1}: reading vs. video viewing. Sensitivity (AUC drop) with refit-bootstrap 95\% interval; bold: interval excludes zero.}
\resizebox{\linewidth}{!}{
\begin{tabular}{lrllllllll}
\toprule
Descriptor & AUC & order & dur.\ coupl. & dur.\ marg. & rotation & jit.\ 0.25$u$ & jit.\ 0.5$u$ & jit.\ 1$u$ & jit.\ 2$u$ \\
\midrule
Count & 0.969 & 0 (ex.) & 0 (ex.) & 0 (ex.) & 0 (ex.) & 0 (ex.) & 0 (ex.) & 0 (ex.) & 0 (ex.) \\
Dispersion & 1.000 & 0 (ex.) & 0 (ex.) & 0 (ex.) & \textbf{+0.021 [+0.01, +0.03]} & +0.000 [$-$0.00, +0.00] & $-$0.000 [$-$0.00, +0.00] & +0.000 [$-$0.00, +0.00] & +0.000 [$-$0.00, +0.00] \\
Classic & 0.969 & \textbf{$-$0.022 [$-$0.03, $-$0.01]} & 0 (ex.) & \textbf{+0.035 [+0.02, +0.05]} & 0 (ex.) & \textbf{$-$0.003 [$-$0.01, $-$0.00]} & \textbf{$-$0.006 [$-$0.01, $-$0.00]} & \textbf{$-$0.007 [$-$0.01, $-$0.00]} & $-$0.004 [$-$0.01, +0.00] \\
Dur.\ trend & 0.957 & +0.003 [$-$0.00, +0.01] & +0.003 [$-$0.00, +0.01] & \textbf{+0.473 [+0.44, +0.50]} & 0 (ex.) & 0 (ex.) & 0 (ex.) & 0 (ex.) & 0 (ex.) \\
RQA $0.5u$ & 0.999 & \textbf{+0.010 [+0.00, +0.02]} & 0 (ex.) & 0 (ex.) & 0 (ex.) & +0.001 [$-$0.00, +0.00] & +0.002 [$-$0.00, +0.00] & \textbf{+0.004 [+0.00, +0.01]} & \textbf{+0.015 [+0.01, +0.02]} \\
RQA $1u$ & 1.000 & \textbf{+0.006 [+0.00, +0.01]} & 0 (ex.) & 0 (ex.) & 0 (ex.) & +0.000 [$-$0.00, +0.00] & +0.001 [$-$0.00, +0.00] & +0.001 [$-$0.00, +0.00] & \textbf{+0.006 [+0.00, +0.01]} \\
Directed & 0.977 & \textbf{$-$0.014 [$-$0.02, $-$0.01]} & 0 (ex.) & 0 (ex.) & \textbf{$-$0.007 [$-$0.01, $-$0.00]} & \textbf{$-$0.011 [$-$0.02, $-$0.00]} & \textbf{$-$0.015 [$-$0.02, $-$0.01]} & \textbf{$-$0.016 [$-$0.02, $-$0.01]} & \textbf{$-$0.013 [$-$0.02, $-$0.01]} \\
FAG lattice & 0.988 & 0 (ex.) & $-$0.001 [$-$0.00, +0.00] & +0.004 [$-$0.00, +0.01] & +0.002 [$-$0.00, +0.01] & $-$0.000 [$-$0.01, +0.00] & +0.001 [$-$0.00, +0.01] & \textbf{+0.009 [+0.00, +0.02]} & \textbf{+0.049 [+0.04, +0.06]} \\
\bottomrule
\end{tabular}}
\end{table}

\begin{table}[!htbp]\centering\scriptsize
\caption{Profile of \texttt{gb\_vd1\_blg}: video viewing vs. ball-catching game. Sensitivity (AUC drop) with refit-bootstrap 95\% interval; bold: interval excludes zero.}
\resizebox{\linewidth}{!}{\begin{tabular}{lrllllllll}
\toprule
Descriptor & AUC & order & dur.\ coupl. & dur.\ marg. & rotation & jit.\ 0.25$u$ & jit.\ 0.5$u$ & jit.\ 1$u$ & jit.\ 2$u$ \\
\midrule
Count & 0.690 & 0 (ex.) & 0 (ex.) & 0 (ex.) & 0 (ex.) & 0 (ex.) & 0 (ex.) & 0 (ex.) & 0 (ex.) \\
Dispersion & 1.000 & 0 (ex.) & 0 (ex.) & 0 (ex.) & +0.000 [$-$0.00, +0.00] & +0.000 [$-$0.00, +0.00] & +0.000 [$-$0.00, +0.00] & $-$0.000 [$-$0.00, +0.00] & +0.000 [$-$0.00, +0.00] \\
Classic & 0.886 & \textbf{$-$0.114 [$-$0.14, $-$0.09]} & 0 (ex.) & \textbf{+0.033 [+0.02, +0.05]} & 0 (ex.) & \textbf{$-$0.003 [$-$0.00, $-$0.00]} & \textbf{$-$0.007 [$-$0.01, $-$0.00]} & \textbf{$-$0.014 [$-$0.02, $-$0.01]} & \textbf{$-$0.014 [$-$0.02, $-$0.01]} \\
Dur.\ trend & 0.752 & +0.011 [$-$0.00, +0.02] & +0.011 [$-$0.00, +0.02] & \textbf{+0.252 [+0.20, +0.30]} & 0 (ex.) & 0 (ex.) & 0 (ex.) & 0 (ex.) & 0 (ex.) \\
RQA $0.5u$ & 0.999 & \textbf{+0.013 [+0.01, +0.02]} & 0 (ex.) & 0 (ex.) & 0 (ex.) & $-$0.000 [$-$0.00, +0.00] & $-$0.000 [$-$0.00, +0.00] & $-$0.000 [$-$0.00, +0.00] & \textbf{+0.003 [+0.00, +0.01]} \\
RQA $1u$ & 1.000 & \textbf{+0.001 [+0.00, +0.00]} & 0 (ex.) & 0 (ex.) & 0 (ex.) & $-$0.000 [$-$0.00, +0.00] & $-$0.000 [$-$0.00, +0.00] & +0.000 [$-$0.00, +0.00] & +0.000 [$-$0.00, +0.00] \\
Directed & 1.000 & \textbf{+0.005 [+0.00, +0.01]} & 0 (ex.) & 0 (ex.) & +0.000 [$-$0.00, +0.00] & +0.000 [$-$0.00, +0.00] & +0.000 [$-$0.00, +0.00] & +0.000 [$-$0.00, +0.00] & \textbf{+0.002 [+0.00, +0.00]} \\
FAG lattice & 0.957 & 0 (ex.) & +0.000 [$-$0.00, +0.00] & \textbf{$-$0.008 [$-$0.01, $-$0.00]} & \textbf{+0.029 [+0.02, +0.04]} & +0.007 [$-$0.00, +0.02] & +0.010 [$-$0.00, +0.02] & \textbf{+0.019 [+0.00, +0.04]} & \textbf{+0.027 [+0.01, +0.04]} \\
\bottomrule
\end{tabular}}
\end{table}

\begin{table}[!htbp]\centering\scriptsize
\caption{Profile of \texttt{gb\_vd1\_vd2}: video segment 1 vs. 2. Sensitivity (AUC drop) with refit-bootstrap 95\% interval; bold: interval excludes zero.}
\resizebox{\linewidth}{!}{\begin{tabular}{lrllllllll}
\toprule
Descriptor & AUC & order & dur.\ coupl. & dur.\ marg. & rotation & jit.\ 0.25$u$ & jit.\ 0.5$u$ & jit.\ 1$u$ & jit.\ 2$u$ \\
\midrule
Count & 0.634 & 0 (ex.) & 0 (ex.) & 0 (ex.) & 0 (ex.) & 0 (ex.) & 0 (ex.) & 0 (ex.) & 0 (ex.) \\
Dispersion & 0.707 & 0 (ex.) & 0 (ex.) & 0 (ex.) & \textbf{+0.098 [+0.07, +0.13]} & +0.004 [$-$0.00, +0.01] & \textbf{+0.012 [+0.00, +0.02]} & \textbf{+0.045 [+0.03, +0.06]} & \textbf{+0.116 [+0.08, +0.15]} \\
Classic & 0.640 & +0.007 [$-$0.00, +0.01] & 0 (ex.) & \textbf{+0.124 [+0.10, +0.15]} & 0 (ex.) & +0.001 [$-$0.00, +0.00] & \textbf{+0.004 [+0.00, +0.01]} & +0.005 [$-$0.00, +0.01] & +0.005 [$-$0.00, +0.01] \\
Dur.\ trend & 0.790 & \textbf{+0.157 [+0.13, +0.19]} & \textbf{+0.158 [+0.13, +0.19]} & \textbf{+0.286 [+0.23, +0.34]} & 0 (ex.) & 0 (ex.) & 0 (ex.) & 0 (ex.) & 0 (ex.) \\
RQA $0.5u$ & 0.677 & \textbf{+0.035 [+0.01, +0.06]} & 0 (ex.) & 0 (ex.) & 0 (ex.) & $-$0.006 [$-$0.03, +0.01] & +0.002 [$-$0.03, +0.03] & \textbf{+0.046 [+0.00, +0.09]} & \textbf{+0.108 [+0.07, +0.14]} \\
RQA $1u$ & 0.719 & \textbf{+0.073 [+0.04, +0.10]} & 0 (ex.) & 0 (ex.) & 0 (ex.) & +0.007 [$-$0.01, +0.02] & +0.013 [$-$0.01, +0.03] & \textbf{+0.043 [+0.01, +0.08]} & \textbf{+0.118 [+0.08, +0.16]} \\
Directed & 0.700 & \textbf{+0.040 [+0.02, +0.06]} & 0 (ex.) & 0 (ex.) & $-$0.010 [$-$0.03, +0.01] & $-$0.005 [$-$0.03, +0.02] & +0.013 [$-$0.01, +0.04] & \textbf{+0.042 [+0.01, +0.08]} & \textbf{+0.069 [+0.04, +0.10]} \\
FAG lattice & 0.664 & 0 (ex.) & \textbf{$-$0.015 [$-$0.03, $-$0.00]} & \textbf{$-$0.020 [$-$0.03, $-$0.01]} & $-$0.012 [$-$0.05, +0.02] & $-$0.015 [$-$0.05, +0.02] & $-$0.006 [$-$0.05, +0.04] & +0.017 [$-$0.03, +0.06] & \textbf{+0.055 [+0.01, +0.10]} \\
\bottomrule
\end{tabular}}
\end{table}

\begin{table}[!htbp]\centering\scriptsize
\caption{Profile of \texttt{gb\_vd1\_halves}: video 1, first vs. second half. Sensitivity (AUC drop) with refit-bootstrap 95\% interval; bold: interval excludes zero.}
\resizebox{\linewidth}{!}{\begin{tabular}{lrllllllll}
\toprule
Descriptor & AUC & order & dur.\ coupl. & dur.\ marg. & rotation & jit.\ 0.25$u$ & jit.\ 0.5$u$ & jit.\ 1$u$ & jit.\ 2$u$ \\
\midrule
Count & 0.500 & 0 (ex.) & 0 (ex.) & 0 (ex.) & 0 (ex.) & 0 (ex.) & 0 (ex.) & 0 (ex.) & 0 (ex.) \\
Dispersion & 0.924 & 0 (ex.) & 0 (ex.) & 0 (ex.) & \textbf{+0.249 [+0.22, +0.27]} & +0.001 [$-$0.00, +0.00] & \textbf{+0.007 [+0.00, +0.01]} & \textbf{+0.029 [+0.02, +0.04]} & \textbf{+0.130 [+0.11, +0.15]} \\
Classic & 0.564 & \textbf{$-$0.090 [$-$0.11, $-$0.07]} & 0 (ex.) & +0.008 [$-$0.01, +0.03] & 0 (ex.) & $-$0.001 [$-$0.00, +0.00] & +0.002 [$-$0.00, +0.01] & +0.005 [$-$0.01, +0.02] & +0.015 [$-$0.00, +0.03] \\
Dur.\ trend & 0.590 & \textbf{+0.076 [+0.03, +0.12]} & \textbf{+0.085 [+0.04, +0.13]} & \textbf{+0.082 [+0.04, +0.12]} & 0 (ex.) & 0 (ex.) & 0 (ex.) & 0 (ex.) & 0 (ex.) \\
RQA $0.5u$ & 0.743 & \textbf{+0.046 [+0.02, +0.07]} & 0 (ex.) & 0 (ex.) & 0 (ex.) & +0.012 [$-$0.01, +0.03] & +0.025 [$-$0.00, +0.05] & \textbf{+0.079 [+0.05, +0.11]} & \textbf{+0.143 [+0.10, +0.19]} \\
RQA $1u$ & 0.823 & \textbf{+0.088 [+0.06, +0.11]} & 0 (ex.) & 0 (ex.) & 0 (ex.) & +0.003 [$-$0.01, +0.02] & \textbf{+0.026 [+0.01, +0.05]} & \textbf{+0.088 [+0.05, +0.12]} & \textbf{+0.182 [+0.15, +0.21]} \\
Directed & 0.805 & \textbf{+0.036 [+0.02, +0.05]} & 0 (ex.) & 0 (ex.) & $-$0.007 [$-$0.03, +0.01] & +0.010 [$-$0.01, +0.03] & \textbf{+0.033 [+0.01, +0.05]} & \textbf{+0.090 [+0.06, +0.12]} & \textbf{+0.191 [+0.16, +0.22]} \\
FAG lattice & 0.741 & 0 (ex.) & $-$0.003 [$-$0.01, +0.00] & +0.000 [$-$0.00, +0.01] & +0.023 [$-$0.01, +0.05] & +0.004 [$-$0.02, +0.03] & +0.011 [$-$0.02, +0.05] & +0.037 [$-$0.00, +0.07] & \textbf{+0.088 [+0.05, +0.12]} \\
\bottomrule
\end{tabular}}
\end{table}

\begin{table}[!htbp]\centering\scriptsize
\caption{Profile of \texttt{gb\_tex\_s1s2}: reading, session 1 vs. 2. Sensitivity (AUC drop) with refit-bootstrap 95\% interval; bold: interval excludes zero.}
\resizebox{\linewidth}{!}{\begin{tabular}{lrllllllll}
\toprule
Descriptor & AUC & order & dur.\ coupl. & dur.\ marg. & rotation & jit.\ 0.25$u$ & jit.\ 0.5$u$ & jit.\ 1$u$ & jit.\ 2$u$ \\
\midrule
Count & 0.530 & 0 (ex.) & 0 (ex.) & 0 (ex.) & 0 (ex.) & 0 (ex.) & 0 (ex.) & 0 (ex.) & 0 (ex.) \\
Dispersion & 0.519 & 0 (ex.) & 0 (ex.) & 0 (ex.) & +0.036 [$-$0.05, +0.12] & +0.003 [$-$0.01, +0.01] & $-$0.000 [$-$0.03, +0.03] & +0.011 [$-$0.03, +0.06] & +0.031 [$-$0.01, +0.07] \\
Classic & 0.539 & +0.005 [$-$0.02, +0.03] & 0 (ex.) & +0.031 [$-$0.04, +0.10] & 0 (ex.) & +0.001 [$-$0.00, +0.01] & $-$0.000 [$-$0.01, +0.01] & $-$0.005 [$-$0.02, +0.02] & $-$0.001 [$-$0.03, +0.03] \\
Dur.\ trend & 0.549 & +0.014 [$-$0.04, +0.07] & +0.016 [$-$0.03, +0.06] & +0.036 [$-$0.05, +0.12] & 0 (ex.) & 0 (ex.) & 0 (ex.) & 0 (ex.) & 0 (ex.) \\
RQA $0.5u$ & 0.576 & \textbf{+0.093 [+0.01, +0.17]} & 0 (ex.) & 0 (ex.) & 0 (ex.) & +0.037 [$-$0.05, +0.12] & \textbf{+0.071 [+0.02, +0.12]} & +0.072 [$-$0.01, +0.15] & +0.062 [$-$0.03, +0.16] \\
RQA $1u$ & 0.521 & +0.047 [$-$0.03, +0.12] & 0 (ex.) & 0 (ex.) & 0 (ex.) & $-$0.007 [$-$0.05, +0.04] & +0.017 [$-$0.06, +0.09] & +0.028 [$-$0.06, +0.12] & +0.007 [$-$0.08, +0.09] \\
Directed & 0.531 & +0.028 [$-$0.05, +0.11] & 0 (ex.) & 0 (ex.) & +0.007 [$-$0.06, +0.07] & +0.003 [$-$0.06, +0.06] & +0.020 [$-$0.03, +0.07] & +0.038 [$-$0.02, +0.10] & +0.039 [$-$0.03, +0.11] \\
FAG lattice & 0.591 & 0 (ex.) & +0.010 [$-$0.01, +0.03] & +0.014 [$-$0.01, +0.04] & +0.042 [$-$0.03, +0.11] & +0.036 [$-$0.04, +0.11] & +0.056 [$-$0.03, +0.14] & +0.067 [$-$0.02, +0.16] & +0.081 [$-$0.01, +0.17] \\
\bottomrule
\end{tabular}}
\end{table}

\begin{table}[!htbp]\centering\scriptsize
\caption{Profile of \texttt{sb\_native}: native vs. non-native English readers, reading pages. Sensitivity (AUC drop) with refit-bootstrap 95\% interval; bold: interval excludes zero.}
\resizebox{\linewidth}{!}{\begin{tabular}{lrllllllll}
\toprule
Descriptor & AUC & order & dur.\ coupl. & dur.\ marg. & rotation & jit.\ 0.25$u$ & jit.\ 0.5$u$ & jit.\ 1$u$ & jit.\ 2$u$ \\
\midrule
Count & 0.624 & 0 (ex.) & 0 (ex.) & 0 (ex.) & 0 (ex.) & 0 (ex.) & 0 (ex.) & 0 (ex.) & 0 (ex.) \\
Dispersion & 0.545 & 0 (ex.) & 0 (ex.) & 0 (ex.) & +0.005 [$-$0.06, +0.07] & +0.000 [$-$0.01, +0.01] & +0.000 [$-$0.02, +0.02] & $-$0.003 [$-$0.04, +0.03] & +0.006 [$-$0.06, +0.07] \\
Classic & 0.689 & +0.037 [$-$0.08, +0.15] & 0 (ex.) & +0.017 [$-$0.07, +0.10] & 0 (ex.) & +0.001 [$-$0.00, +0.00] & +0.004 [$-$0.00, +0.01] & +0.012 [$-$0.01, +0.03] & +0.022 [$-$0.04, +0.08] \\
Dur.\ trend & 0.654 & +0.011 [$-$0.04, +0.06] & +0.012 [$-$0.03, +0.06] & \textbf{+0.144 [+0.00, +0.29]} & 0 (ex.) & 0 (ex.) & 0 (ex.) & 0 (ex.) & 0 (ex.) \\
RQA $0.5u$ & 0.614 & +0.074 [$-$0.01, +0.16] & 0 (ex.) & 0 (ex.) & 0 (ex.) & +0.012 [$-$0.02, +0.04] & +0.029 [$-$0.01, +0.07] & \textbf{+0.079 [+0.01, +0.15]} & \textbf{+0.088 [+0.01, +0.16]} \\
RQA $1u$ & 0.669 & +0.105 [$-$0.00, +0.21] & 0 (ex.) & 0 (ex.) & 0 (ex.) & $-$0.006 [$-$0.03, +0.02] & +0.015 [$-$0.02, +0.05] & \textbf{+0.086 [+0.02, +0.16]} & \textbf{+0.140 [+0.05, +0.23]} \\
Directed & 0.614 & $-$0.030 [$-$0.09, +0.03] & 0 (ex.) & 0 (ex.) & $-$0.004 [$-$0.04, +0.03] & $-$0.011 [$-$0.04, +0.02] & $-$0.015 [$-$0.06, +0.03] & $-$0.014 [$-$0.06, +0.04] & $-$0.004 [$-$0.07, +0.06] \\
FAG lattice & 0.587 & 0 (ex.) & $-$0.000 [$-$0.02, +0.02] & +0.008 [$-$0.02, +0.03] & $-$0.002 [$-$0.04, +0.03] & $-$0.004 [$-$0.03, +0.02] & $-$0.011 [$-$0.06, +0.03] & $-$0.018 [$-$0.07, +0.04] & $-$0.022 [$-$0.09, +0.04] \\
\bottomrule
\end{tabular}}
\end{table}

\begin{table}[!htbp]\centering\scriptsize
\caption{Profile of \texttt{sb\_compr}: high vs. low passage comprehension, reading pages. Sensitivity (AUC drop) with refit-bootstrap 95\% interval; bold: interval excludes zero.}
\resizebox{\linewidth}{!}{\begin{tabular}{lrllllllll}
\toprule
Descriptor & AUC & order & dur.\ coupl. & dur.\ marg. & rotation & jit.\ 0.25$u$ & jit.\ 0.5$u$ & jit.\ 1$u$ & jit.\ 2$u$ \\
\midrule
Count & 0.554 & 0 (ex.) & 0 (ex.) & 0 (ex.) & 0 (ex.) & 0 (ex.) & 0 (ex.) & 0 (ex.) & 0 (ex.) \\
Dispersion & 0.413 & 0 (ex.) & 0 (ex.) & 0 (ex.) & $-$0.047 [$-$0.12, +0.03] & $-$0.004 [$-$0.02, +0.01] & $-$0.007 [$-$0.03, +0.01] & \textbf{$-$0.035 [$-$0.06, $-$0.01]} & $-$0.041 [$-$0.11, +0.03] \\
Classic & 0.587 & +0.010 [$-$0.07, +0.09] & 0 (ex.) & +0.026 [$-$0.09, +0.14] & 0 (ex.) & +0.000 [$-$0.00, +0.00] & +0.001 [$-$0.01, +0.01] & +0.000 [$-$0.01, +0.01] & +0.003 [$-$0.04, +0.05] \\
Dur.\ trend & 0.575 & $-$0.006 [$-$0.04, +0.03] & $-$0.007 [$-$0.03, +0.02] & +0.070 [$-$0.05, +0.19] & 0 (ex.) & 0 (ex.) & 0 (ex.) & 0 (ex.) & 0 (ex.) \\
RQA $0.5u$ & 0.497 & +0.011 [$-$0.12, +0.14] & 0 (ex.) & 0 (ex.) & 0 (ex.) & +0.009 [$-$0.06, +0.07] & $-$0.003 [$-$0.10, +0.09] & +0.001 [$-$0.08, +0.08] & $-$0.005 [$-$0.11, +0.09] \\
RQA $1u$ & 0.529 & +0.050 [$-$0.04, +0.14] & 0 (ex.) & 0 (ex.) & 0 (ex.) & $-$0.017 [$-$0.06, +0.03] & $-$0.014 [$-$0.07, +0.04] & +0.006 [$-$0.05, +0.07] & +0.032 [$-$0.08, +0.15] \\
Directed & 0.533 & +0.004 [$-$0.07, +0.08] & 0 (ex.) & 0 (ex.) & $-$0.010 [$-$0.10, +0.07] & +0.013 [$-$0.05, +0.07] & $-$0.004 [$-$0.08, +0.07] & +0.003 [$-$0.07, +0.08] & +0.010 [$-$0.07, +0.09] \\
FAG lattice & 0.506 & 0 (ex.) & $-$0.007 [$-$0.04, +0.02] & $-$0.002 [$-$0.03, +0.03] & $-$0.048 [$-$0.10, +0.00] & $-$0.032 [$-$0.10, +0.04] & $-$0.035 [$-$0.11, +0.04] & $-$0.038 [$-$0.13, +0.05] & $-$0.011 [$-$0.10, +0.08] \\
\bottomrule
\end{tabular}}
\end{table}

\begin{table}[!htbp]\centering\scriptsize
\caption{Profile of \texttt{em\_program}: Rectangle vs. Vehicle program. Sensitivity (AUC drop) with refit-bootstrap 95\% interval; bold: interval excludes zero.}
\resizebox{\linewidth}{!}{\begin{tabular}{lrllllllll}
\toprule
Descriptor & AUC & order & dur.\ coupl. & dur.\ marg. & rotation & jit.\ 0.25$u$ & jit.\ 0.5$u$ & jit.\ 1$u$ & jit.\ 2$u$ \\
\midrule
Count & 0.561 & 0 (ex.) & 0 (ex.) & 0 (ex.) & 0 (ex.) & 0 (ex.) & 0 (ex.) & 0 (ex.) & 0 (ex.) \\
Dispersion & 0.957 & 0 (ex.) & 0 (ex.) & 0 (ex.) & \textbf{+0.139 [+0.08, +0.20]} & $-$0.001 [$-$0.01, +0.00] & $-$0.001 [$-$0.01, +0.01] & +0.001 [$-$0.01, +0.01] & +0.002 [$-$0.01, +0.02] \\
Classic & 0.539 & \textbf{$-$0.249 [$-$0.34, $-$0.16]} & 0 (ex.) & +0.032 [$-$0.08, +0.15] & 0 (ex.) & $-$0.000 [$-$0.02, +0.02] & +0.003 [$-$0.03, +0.04] & $-$0.004 [$-$0.06, +0.05] & +0.008 [$-$0.10, +0.11] \\
Dur.\ trend & 0.540 & +0.017 [$-$0.13, +0.16] & +0.038 [$-$0.12, +0.20] & +0.055 [$-$0.07, +0.18] & 0 (ex.) & 0 (ex.) & 0 (ex.) & 0 (ex.) & 0 (ex.) \\
RQA $0.5u$ & 0.580 & +0.036 [$-$0.07, +0.14] & 0 (ex.) & 0 (ex.) & 0 (ex.) & +0.004 [$-$0.09, +0.09] & +0.061 [$-$0.08, +0.21] & +0.068 [$-$0.12, +0.25] & +0.000 [$-$0.17, +0.17] \\
RQA $1u$ & 0.547 & +0.003 [$-$0.15, +0.16] & 0 (ex.) & 0 (ex.) & 0 (ex.) & $-$0.010 [$-$0.10, +0.08] & +0.003 [$-$0.13, +0.13] & $-$0.046 [$-$0.22, +0.13] & $-$0.067 [$-$0.26, +0.12] \\
Directed & 0.655 & $-$0.003 [$-$0.08, +0.07] & 0 (ex.) & 0 (ex.) & +0.090 [$-$0.00, +0.18] & +0.112 [$-$0.02, +0.24] & +0.141 [$-$0.01, +0.29] & +0.035 [$-$0.08, +0.15] & +0.044 [$-$0.10, +0.19] \\
\bottomrule
\end{tabular}}
\end{table}

\begin{table}[!htbp]\centering\scriptsize
\caption{Profile of \texttt{em\_halves}: first vs. second half of a program reading. Sensitivity (AUC drop) with refit-bootstrap 95\% interval; bold: interval excludes zero.}
\resizebox{\linewidth}{!}{\begin{tabular}{lrllllllll}
\toprule
Descriptor & AUC & order & dur.\ coupl. & dur.\ marg. & rotation & jit.\ 0.25$u$ & jit.\ 0.5$u$ & jit.\ 1$u$ & jit.\ 2$u$ \\
\midrule
Count & 0.500 & 0 (ex.) & 0 (ex.) & 0 (ex.) & 0 (ex.) & 0 (ex.) & 0 (ex.) & 0 (ex.) & 0 (ex.) \\
Dispersion & 0.616 & 0 (ex.) & 0 (ex.) & 0 (ex.) & +0.012 [$-$0.05, +0.07] & $-$0.001 [$-$0.01, +0.01] & +0.000 [$-$0.01, +0.01] & +0.001 [$-$0.02, +0.02] & +0.008 [$-$0.02, +0.04] \\
Classic & 0.776 & \textbf{+0.172 [+0.11, +0.23]} & 0 (ex.) & +0.004 [$-$0.01, +0.02] & 0 (ex.) & $-$0.004 [$-$0.01, +0.00] & $-$0.003 [$-$0.02, +0.01] & $-$0.002 [$-$0.03, +0.02] & +0.043 [$-$0.00, +0.09] \\
Dur.\ trend & 0.582 & +0.048 [$-$0.04, +0.14] & +0.036 [$-$0.08, +0.15] & \textbf{+0.086 [+0.00, +0.17]} & 0 (ex.) & 0 (ex.) & 0 (ex.) & 0 (ex.) & 0 (ex.) \\
RQA $0.5u$ & 0.613 & +0.058 [$-$0.04, +0.15] & 0 (ex.) & 0 (ex.) & 0 (ex.) & $-$0.036 [$-$0.09, +0.02] & $-$0.050 [$-$0.11, +0.01] & \textbf{$-$0.085 [$-$0.15, $-$0.02]} & $-$0.053 [$-$0.11, +0.01] \\
RQA $1u$ & 0.680 & +0.081 [$-$0.00, +0.16] & 0 (ex.) & 0 (ex.) & 0 (ex.) & $-$0.006 [$-$0.02, +0.01] & $-$0.018 [$-$0.05, +0.01] & \textbf{$-$0.052 [$-$0.10, $-$0.01]} & $-$0.043 [$-$0.09, +0.01] \\
Directed & 0.634 & +0.033 [$-$0.04, +0.10] & 0 (ex.) & 0 (ex.) & $-$0.004 [$-$0.06, +0.05] & $-$0.010 [$-$0.07, +0.05] & $-$0.010 [$-$0.07, +0.05] & +0.002 [$-$0.07, +0.08] & +0.059 [$-$0.03, +0.15] \\
\bottomrule
\end{tabular}}
\end{table}

\begin{table}[!htbp]\centering\scriptsize
\caption{Profile of \texttt{em\_expertise}: high vs. low/no self-rated programming expertise. Sensitivity (AUC drop) with refit-bootstrap 95\% interval; bold: interval excludes zero.}
\resizebox{\linewidth}{!}{\begin{tabular}{lrllllllll}
\toprule
Descriptor & AUC & order & dur.\ coupl. & dur.\ marg. & rotation & jit.\ 0.25$u$ & jit.\ 0.5$u$ & jit.\ 1$u$ & jit.\ 2$u$ \\
\midrule
Count & 0.472 & 0 (ex.) & 0 (ex.) & 0 (ex.) & 0 (ex.) & 0 (ex.) & 0 (ex.) & 0 (ex.) & 0 (ex.) \\
Dispersion & 0.469 & 0 (ex.) & 0 (ex.) & 0 (ex.) & $-$0.028 [$-$0.21, +0.15] & +0.006 [$-$0.03, +0.04] & +0.004 [$-$0.05, +0.06] & +0.007 [$-$0.06, +0.08] & $-$0.004 [$-$0.10, +0.09] \\
Classic & 0.438 & $-$0.032 [$-$0.28, +0.21] & 0 (ex.) & $-$0.023 [$-$0.34, +0.29] & 0 (ex.) & $-$0.002 [$-$0.04, +0.03] & $-$0.008 [$-$0.05, +0.03] & $-$0.043 [$-$0.10, +0.01] & $-$0.031 [$-$0.16, +0.09] \\
Dur.\ trend & 0.541 & +0.046 [$-$0.13, +0.23] & +0.070 [$-$0.23, +0.37] & +0.031 [$-$0.27, +0.33] & 0 (ex.) & 0 (ex.) & 0 (ex.) & 0 (ex.) & 0 (ex.) \\
RQA $0.5u$ & 0.419 & $-$0.016 [$-$0.37, +0.34] & 0 (ex.) & 0 (ex.) & 0 (ex.) & $-$0.022 [$-$0.21, +0.17] & $-$0.033 [$-$0.11, +0.04] & +0.068 [$-$0.13, +0.27] & +0.056 [$-$0.21, +0.32] \\
RQA $1u$ & 0.416 & $-$0.037 [$-$0.37, +0.30] & 0 (ex.) & 0 (ex.) & 0 (ex.) & $-$0.017 [$-$0.13, +0.09] & +0.016 [$-$0.24, +0.28] & +0.028 [$-$0.21, +0.27] & +0.090 [$-$0.24, +0.42] \\
Directed & 0.366 & +0.040 [$-$0.12, +0.20] & 0 (ex.) & 0 (ex.) & $-$0.018 [$-$0.10, +0.06] & +0.044 [$-$0.17, +0.26] & +0.013 [$-$0.25, +0.28] & $-$0.086 [$-$0.31, +0.14] & $-$0.083 [$-$0.32, +0.15] \\
\bottomrule
\end{tabular}}
\end{table}

\begin{table}[!htbp]\centering\scriptsize
\caption{Profile of \texttt{em\_correct}: correct vs. incorrect answer, both programs pooled (confounded). Sensitivity (AUC drop) with refit-bootstrap 95\% interval; bold: interval excludes zero.}
\resizebox{\linewidth}{!}{\begin{tabular}{lrllllllll}
\toprule
Descriptor & AUC & order & dur.\ coupl. & dur.\ marg. & rotation & jit.\ 0.25$u$ & jit.\ 0.5$u$ & jit.\ 1$u$ & jit.\ 2$u$ \\
\midrule
Count & 0.495 & 0 (ex.) & 0 (ex.) & 0 (ex.) & 0 (ex.) & 0 (ex.) & 0 (ex.) & 0 (ex.) & 0 (ex.) \\
Dispersion & 0.716 & 0 (ex.) & 0 (ex.) & 0 (ex.) & \textbf{+0.096 [+0.01, +0.18]} & +0.000 [$-$0.01, +0.01] & +0.003 [$-$0.01, +0.01] & $-$0.001 [$-$0.01, +0.01] & +0.005 [$-$0.03, +0.04] \\
Classic & 0.501 & $-$0.123 [$-$0.27, +0.03] & 0 (ex.) & +0.054 [$-$0.07, +0.18] & 0 (ex.) & $-$0.000 [$-$0.02, +0.02] & $-$0.000 [$-$0.03, +0.03] & $-$0.004 [$-$0.06, +0.05] & $-$0.002 [$-$0.07, +0.07] \\
Dur.\ trend & 0.511 & $-$0.013 [$-$0.17, +0.15] & $-$0.023 [$-$0.17, +0.13] & +0.046 [$-$0.09, +0.18] & 0 (ex.) & 0 (ex.) & 0 (ex.) & 0 (ex.) & 0 (ex.) \\
RQA $0.5u$ & 0.571 & +0.022 [$-$0.10, +0.14] & 0 (ex.) & 0 (ex.) & 0 (ex.) & +0.040 [$-$0.05, +0.13] & +0.056 [$-$0.05, +0.16] & +0.097 [$-$0.06, +0.25] & +0.061 [$-$0.12, +0.24] \\
RQA $1u$ & 0.525 & $-$0.031 [$-$0.19, +0.13] & 0 (ex.) & 0 (ex.) & 0 (ex.) & +0.018 [$-$0.05, +0.09] & +0.069 [$-$0.04, +0.18] & +0.062 [$-$0.08, +0.20] & +0.009 [$-$0.18, +0.20] \\
Directed & 0.580 & $-$0.021 [$-$0.11, +0.07] & 0 (ex.) & 0 (ex.) & +0.043 [$-$0.05, +0.14] & +0.053 [$-$0.04, +0.15] & +0.103 [$-$0.02, +0.23] & +0.095 [$-$0.05, +0.24] & +0.058 [$-$0.12, +0.23] \\
\bottomrule
\end{tabular}}
\end{table}

\begin{table}[!htbp]\centering\scriptsize
\caption{Profile of \texttt{em\_correctrect}: correct vs. incorrect answer, Rectangle program only. Sensitivity (AUC drop) with refit-bootstrap 95\% interval; bold: interval excludes zero.}
\resizebox{\linewidth}{!}{\begin{tabular}{lrllllllll}
\toprule
Descriptor & AUC & order & dur.\ coupl. & dur.\ marg. & rotation & jit.\ 0.25$u$ & jit.\ 0.5$u$ & jit.\ 1$u$ & jit.\ 2$u$ \\
\midrule
Count & 0.589 & 0 (ex.) & 0 (ex.) & 0 (ex.) & 0 (ex.) & 0 (ex.) & 0 (ex.) & 0 (ex.) & 0 (ex.) \\
Dispersion & 0.472 & 0 (ex.) & 0 (ex.) & 0 (ex.) & $-$0.062 [$-$0.25, +0.13] & +0.002 [$-$0.02, +0.02] & +0.006 [$-$0.03, +0.04] & +0.034 [$-$0.02, +0.09] & +0.073 [$-$0.02, +0.17] \\
Classic & 0.523 & $-$0.002 [$-$0.19, +0.19] & 0 (ex.) & +0.048 [$-$0.19, +0.28] & 0 (ex.) & +0.004 [$-$0.03, +0.04] & +0.008 [$-$0.05, +0.06] & $-$0.006 [$-$0.08, +0.06] & +0.003 [$-$0.08, +0.09] \\
Dur.\ trend & 0.507 & $-$0.069 [$-$0.23, +0.09] & $-$0.077 [$-$0.24, +0.08] & $-$0.037 [$-$0.29, +0.22] & 0 (ex.) & 0 (ex.) & 0 (ex.) & 0 (ex.) & 0 (ex.) \\
RQA $0.5u$ & 0.451 & $-$0.054 [$-$0.38, +0.27] & 0 (ex.) & 0 (ex.) & 0 (ex.) & $-$0.046 [$-$0.25, +0.16] & $-$0.072 [$-$0.22, +0.08] & $-$0.004 [$-$0.26, +0.25] & $-$0.041 [$-$0.28, +0.20] \\
RQA $1u$ & 0.462 & $-$0.065 [$-$0.38, +0.25] & 0 (ex.) & 0 (ex.) & 0 (ex.) & $-$0.015 [$-$0.15, +0.12] & $-$0.026 [$-$0.18, +0.13] & $-$0.046 [$-$0.24, +0.15] & $-$0.048 [$-$0.29, +0.20] \\
Directed & 0.533 & +0.006 [$-$0.22, +0.23] & 0 (ex.) & 0 (ex.) & $-$0.013 [$-$0.19, +0.17] & +0.000 [$-$0.18, +0.19] & +0.014 [$-$0.19, +0.22] & $-$0.000 [$-$0.22, +0.22] & +0.018 [$-$0.19, +0.23] \\
\bottomrule
\end{tabular}}
\end{table}

\begin{table}[!htbp]\centering\scriptsize
\caption{Profile of \texttt{em\_correctveh}: correct vs. incorrect answer, Vehicle program only. Sensitivity (AUC drop) with refit-bootstrap 95\% interval; bold: interval excludes zero.}
\resizebox{\linewidth}{!}{\begin{tabular}{lrllllllll}
\toprule
Descriptor & AUC & order & dur.\ coupl. & dur.\ marg. & rotation & jit.\ 0.25$u$ & jit.\ 0.5$u$ & jit.\ 1$u$ & jit.\ 2$u$ \\
\midrule
Count & 0.444 & 0 (ex.) & 0 (ex.) & 0 (ex.) & 0 (ex.) & 0 (ex.) & 0 (ex.) & 0 (ex.) & 0 (ex.) \\
Dispersion & 0.476 & 0 (ex.) & 0 (ex.) & 0 (ex.) & +0.091 [$-$0.20, +0.38] & +0.006 [$-$0.02, +0.03] & +0.007 [$-$0.04, +0.06] & +0.015 [$-$0.04, +0.07] & +0.022 [$-$0.10, +0.15] \\
Classic & 0.500 & +0.070 [$-$0.15, +0.29] & 0 (ex.) & +0.025 [$-$0.20, +0.25] & 0 (ex.) & +0.000 [$-$0.04, +0.04] & +0.001 [$-$0.06, +0.06] & +0.003 [$-$0.05, +0.06] & +0.017 [$-$0.10, +0.14] \\
Dur.\ trend & 0.521 & +0.012 [$-$0.22, +0.24] & $-$0.024 [$-$0.26, +0.21] & +0.002 [$-$0.30, +0.31] & 0 (ex.) & 0 (ex.) & 0 (ex.) & 0 (ex.) & 0 (ex.) \\
RQA $0.5u$ & 0.518 & $-$0.026 [$-$0.22, +0.17] & 0 (ex.) & 0 (ex.) & 0 (ex.) & +0.034 [$-$0.15, +0.21] & +0.026 [$-$0.23, +0.28] & +0.104 [$-$0.14, +0.35] & +0.098 [$-$0.11, +0.31] \\
RQA $1u$ & 0.559 & +0.036 [$-$0.24, +0.31] & 0 (ex.) & 0 (ex.) & 0 (ex.) & $-$0.004 [$-$0.11, +0.10] & +0.093 [$-$0.10, +0.29] & +0.106 [$-$0.09, +0.30] & +0.140 [$-$0.07, +0.35] \\
Directed & 0.618 & +0.134 [$-$0.06, +0.33] & 0 (ex.) & 0 (ex.) & +0.080 [$-$0.14, +0.31] & +0.097 [$-$0.11, +0.30] & +0.120 [$-$0.05, +0.29] & +0.147 [$-$0.07, +0.37] & +0.088 [$-$0.15, +0.33] \\
\bottomrule
\end{tabular}}
\end{table}

%% file: tab_auc_pooled.tex
\begin{tabular}{lrrrrrrrrrr}
\toprule
Problem & Count & Dispersion & Classic & Dur.\ trend & RQA $0.5u$ & RQA $1u$ & Directed & FAG lattice & FAG AOI & FAG AOI pad. \\
\midrule
\texttt{gb\_tex\_vd1} & 0.968 (.005) & 1.000 (.005) & 0.968 (.005) & 0.956 (.005) & 0.999 (.005) & 1.000 (.005) & 0.977 (.005) & 0.989 (.005) & -- & -- \\
\texttt{gb\_vd1\_blg} & 0.691 (.005) & 1.000 (.005) & 0.888 (.005) & 0.751 (.005) & 0.999 (.005) & 1.000 (.005) & 1.000 (.005) & 0.957 (.005) & -- & -- \\
\texttt{gb\_vd1\_vd2} & 0.633 (.005) & 0.703 (.005) & 0.636 (.005) & 0.791 (.005) & 0.672 (.005) & 0.715 (.005) & 0.700 (.005) & 0.666 (.005) & -- & -- \\
\texttt{gb\_vd1\_halves} & 0.500 (1.000) & 0.921 (.005) & 0.561 (.005) & 0.590 (.005) & 0.742 (.005) & 0.822 (.005) & 0.803 (.005) & 0.739 (.005) & -- & -- \\
\texttt{gb\_tex\_s1s2} & 0.516 (.065) & 0.517 (.224) & 0.537 (.005) & 0.545 (.015) & 0.574 (.005) & 0.515 (.159) & 0.531 (.070) & 0.588 (.005) & -- & -- \\
\texttt{sb\_native} & 0.610 (.010) & 0.530 (.164) & 0.673 (.005) & 0.637 (.015) & 0.611 (.005) & 0.659 (.005) & 0.594 (.025) & 0.575 (.050) & -- & -- \\
\texttt{sb\_compr} & 0.527 (.915) & 0.402 (.990) & 0.595 (.950) & 0.575 (.910) & 0.485 (.896) & 0.523 (.891) & 0.514 (.935) & 0.473 (.955) & -- & -- \\
\bottomrule
\end{tabular}

%% file: supp_null_means.tex
\begin{tabular}{lrr}
\toprule
Problem & Within-fold AUC & Pooled AUC \\
\midrule
\texttt{gb\_tex\_vd1} & 0.499--0.501 & 0.491--0.496 \\
\texttt{gb\_vd1\_blg} & 0.499--0.502 & 0.494--0.497 \\
\texttt{gb\_vd1\_vd2} & 0.499--0.501 & 0.494--0.499 \\
\texttt{gb\_vd1\_halves} & 0.499--0.501 & 0.496--0.500 \\
\texttt{gb\_tex\_s1s2} & 0.499--0.501 & 0.494--0.498 \\
\texttt{sb\_native} & 0.497--0.504 & 0.476--0.490 \\
\texttt{sb\_compr} & 0.477--0.618 & 0.458--0.611 \\
\texttt{em\_program} & 0.496--0.501 & 0.488--0.500 \\
\texttt{em\_halves} & 0.498--0.502 & 0.487--0.500 \\
\texttt{em\_expertise} & 0.489--0.504 & 0.466--0.492 \\
\texttt{em\_correct} & 0.502--0.710 & 0.487--0.708 \\
\texttt{em\_correctrect} & 0.496--0.505 & 0.476--0.492 \\
\texttt{em\_correctveh} & 0.496--0.503 & 0.475--0.491 \\
\bottomrule
\end{tabular}

%% file: supp_ledger.tex
\begin{longtable}{lllllllllrc}
\caption{Evidence ledger of the 89 interpreted cells. Primary: sensitivity with the within-fold AUC and class-balanced folds ($R = 6$, $B = 60$, seed 42), with its 95\% refit-bootstrap interval (excluding zero by selection). Pooled AUC: the same cell with the pooled AUC. Partitions: range of the point estimate over 20 fold partitions. Long runs: range of the point estimate over three runs with $R = 20$ and $B = 300$ (seeds 42, 43, 44), and the number of those runs whose interval excluded zero. Trees: gradient-boosted trees ($R = 6$, $B = 60$, seed 42). Norm.: sensitivity divided by the observed AUC minus the mean of its permutation distribution, under the within-fold and the pooled AUC. Robust: sign and exclusion of zero hold in the primary run, the pooled run, every partition and every long run and, for the AOI graph, on the padded graph (Table S23 and Section 5.9). Bold: interval excludes zero.}\\
\toprule
Problem & Descriptor & Transformation & Primary & Pooled AUC & Partitions & Long runs & Trees & Norm. & Norm.\ (pooled) & Robust \\
\midrule
\endfirsthead
\toprule
Problem & Descriptor & Transformation & Primary & Pooled AUC & Partitions & Long runs & Trees & Norm. & Norm.\ (pooled) & Robust \\
\midrule
\endhead
\bottomrule
\endfoot
\texttt{gb\_tex\_vd1} & Classic & duration marginal & +0.035 [+0.021, +0.048] & \textbf{+0.035 [+0.022, +0.048]} & $+0.033$ to $+0.036$ & $+0.034$ to $+0.034$ (3/3) & \textbf{+0.045 [+0.025, +0.065]} & $+0.07$ & $+0.07$ & $\checkmark$ \\
\texttt{gb\_tex\_vd1} & Classic & order & $-$0.022 [$-$0.034, $-$0.010] & \textbf{$-$0.022 [$-$0.034, $-$0.009]} & $-0.024$ to $-0.020$ & $-0.022$ to $-0.022$ (3/3) & \textbf{$-$0.019 [$-$0.033, $-$0.005]} & $-0.05$ & $-0.05$ & $\checkmark$ \\
\texttt{gb\_tex\_vd1} & Classic & jitter $0.25u$ & $-$0.003 [$-$0.005, $-$0.001] & \textbf{$-$0.003 [$-$0.005, $-$0.001]} & $-0.003$ to $-0.002$ & $-0.003$ to $-0.003$ (3/3) & $-$0.002 [$-$0.009, +0.004] & $-0.01$ & $-0.01$ & $\checkmark$ \\
\texttt{gb\_tex\_vd1} & Classic & jitter $0.5u$ & $-$0.006 [$-$0.009, $-$0.002] & \textbf{$-$0.006 [$-$0.008, $-$0.003]} & $-0.007$ to $-0.005$ & $-0.006$ to $-0.005$ (3/3) & $-$0.005 [$-$0.011, +0.001] & $-0.01$ & $-0.01$ & $\checkmark$ \\
\texttt{gb\_tex\_vd1} & Classic & jitter $1u$ & $-$0.007 [$-$0.011, $-$0.003] & \textbf{$-$0.007 [$-$0.011, $-$0.003]} & $-0.009$ to $-0.007$ & $-0.007$ to $-0.007$ (3/3) & $-$0.006 [$-$0.014, +0.003] & $-0.01$ & $-0.02$ & $\checkmark$ \\
\texttt{gb\_tex\_vd1} & Directed & order & $-$0.014 [$-$0.020, $-$0.008] & \textbf{$-$0.014 [$-$0.020, $-$0.008]} & $-0.014$ to $-0.012$ & $-0.015$ to $-0.014$ (3/3) & $-$0.004 [$-$0.011, +0.002] & $-0.03$ & $-0.03$ & $\checkmark$ \\
\texttt{gb\_tex\_vd1} & Directed & jitter $0.25u$ & $-$0.011 [$-$0.018, $-$0.004] & \textbf{$-$0.011 [$-$0.016, $-$0.005]} & $-0.013$ to $-0.011$ & $-0.012$ to $-0.011$ (3/3) & $-$0.004 [$-$0.010, +0.002] & $-0.02$ & $-0.02$ & $\checkmark$ \\
\texttt{gb\_tex\_vd1} & Directed & jitter $0.5u$ & $-$0.015 [$-$0.021, $-$0.009] & \textbf{$-$0.015 [$-$0.020, $-$0.009]} & $-0.015$ to $-0.013$ & $-0.016$ to $-0.015$ (3/3) & $-$0.005 [$-$0.011, +0.001] & $-0.03$ & $-0.03$ & $\checkmark$ \\
\end{longtable}

%% file: supp_se.tex
\begin{tabular}{llrrrrr}
\toprule
Design & Size & Contrasts & SE ratio, median (range) & Coverage at 80\% & at 90\% & at 95\% \\
\midrule
no signal & 65 $\times$ 2 (B) & 8 & 1.66 (1.35--2.04) & 0.959 & 0.988 & 0.997 \\
no signal & 95 $\times$ 20 (B) & 8 & 1.48 (1.23--2.03) & 0.922 & 0.978 & 0.988 \\
no signal & 112 $\times$ 2 (W) & 8 & 1.80 (1.66--2.02) & 0.963 & 0.994 & 1.000 \\
no signal & 60 $\times$ 4 (W) & 8 & 1.80 (1.48--2.22) & 0.966 & 0.987 & 1.000 \\
signal elsewhere & 65 $\times$ 2 (B) & 5 & 2.40 (1.94--3.10) & 0.985 & 0.990 & 1.000 \\
signal elsewhere & 95 $\times$ 20 (B) & 5 & 1.46 (1.28--1.65) & 0.925 & 0.985 & 1.000 \\
signal elsewhere & 112 $\times$ 2 (W) & 5 & 2.35 (2.04--2.82) & 0.990 & 1.000 & 1.000 \\
signal elsewhere & 60 $\times$ 4 (W) & 5 & 2.65 (1.76--3.37) & 0.995 & 1.000 & 1.000 \\
signal elsewhere, no offset & 65 $\times$ 2 (B) & 3 & 2.40 (1.87--2.58) & 1.000 & 1.000 & 1.000 \\
signal elsewhere, no offset & 95 $\times$ 20 (B) & 3 & 1.29 (1.28--1.66) & 0.925 & 0.967 & 0.983 \\
signal elsewhere, no offset & 112 $\times$ 2 (W) & 3 & 1.97 (1.57--2.82) & 0.975 & 1.000 & 1.000 \\
signal elsewhere, no offset & 60 $\times$ 4 (W) & 3 & 2.19 (1.67--3.37) & 0.975 & 1.000 & 1.000 \\
\bottomrule
\end{tabular}

%% file: supp_direct.tex
\begin{tabular}{llllrrrrrr}
\toprule
 & & & & \multicolumn{2}{c}{Coverage} & \multicolumn{2}{c}{Width} & \multicolumn{2}{c}{Excl.\ 0} \\
\cmidrule(lr){5-6}\cmidrule(lr){7-8}\cmidrule(lr){9-10}
Design & Size & Descr. & Contrast & direct & corr. & direct & corr. & direct & corr. \\
\midrule
no signal & 65 $\times$ 2 (B) & rqa & order & 0.985 & 0.990 & 0.367 & 0.434 & 0.015 & 0.010 \\
no signal & 65 $\times$ 2 (B) & rqa & jitter 160 px & 0.995 & 1.000 & 0.419 & 0.505 & 0.005 & 0.000 \\
no signal & 65 $\times$ 2 (B) & rqa & jitter 40 px & 0.985 & 0.990 & 0.417 & 0.498 & 0.015 & 0.010 \\
no signal & 60 $\times$ 4 (W) & classic & order & 0.995 & 0.985 & 0.241 & 0.286 & 0.005 & 0.015 \\
no signal & 60 $\times$ 4 (W) & rqa & order & 0.990 & 0.995 & 0.272 & 0.324 & 0.010 & 0.005 \\
no signal & 60 $\times$ 4 (W) & rqa & jitter 160 px & 0.995 & 1.000 & 0.313 & 0.379 & 0.005 & 0.005 \\
no signal & 60 $\times$ 4 (W) & rqa & jitter 40 px & 0.990 & 1.000 & 0.314 & 0.376 & 0.010 & 0.000 \\
signal elsewhere & 65 $\times$ 2 (B) & rqa & order & 1.000 & 1.000 & 0.125 & 0.161 & 0.000 & 0.000 \\
signal elsewhere & 65 $\times$ 2 (B) & rqa & jitter 40 px & 1.000 & 1.000 & 0.219 & 0.345 & 1.000 & 0.955 \\
signal elsewhere & 60 $\times$ 4 (W) & rqa & order & 1.000 & 1.000 & 0.070 & 0.089 & 0.000 & 0.000 \\
signal elsewhere & 60 $\times$ 4 (W) & rqa & jitter 40 px & 0.995 & 1.000 & 0.147 & 0.238 & 1.000 & 1.000 \\
signal elsewhere, no offset & 60 $\times$ 4 (W) & rqa & order & 1.000 & 1.000 & 0.070 & 0.089 & 0.000 & 0.000 \\
\bottomrule
\end{tabular}

%% file: supp_ext.tex
\begin{longtable}{lllrrrrr}
\caption{Extended simulation designs ($R = 6$, $B = 60$; 40 datasets per row). Target: the reference value (60 datasets with disjoint seeds, $R = 20$) with its Monte Carlo standard error, or zero by exchangeability. Unequal: one to four trials per class and participant (W) or per participant (B); 1:3 and 16:49: class imbalance; heterogeneous: participant offset SD 100 px and participant-specific scaling.}\\
\toprule
Design & Size & Contrast & Target $\theta$ (SE) & Coverage & Excl.\ 0 & Width & Obs.\ AUC \\
\midrule
\endfirsthead
\toprule
Design & Size & Contrast & Target $\theta$ (SE) & Coverage & Excl.\ 0 & Width & Obs.\ AUC \\
\midrule
\endhead
\bottomrule
\endfoot
order hides spread (classic) & 65 $\times$ 2 (B) & duration marginal & +0.006 (0.009) & 0.975 & 0.025 & 0.397 & 0.520 \\
order hides spread (classic) & 65 $\times$ 2 (B) & order & $-$0.499 (0.010) & 1.000 & 1.000 & 0.405 & 0.520 \\
order hides spread (classic) & 60 $\times$ 4 (W) & duration marginal & $-$0.001 (0.005) & 1.000 & 0.000 & 0.296 & 0.509 \\
order hides spread (classic) & 60 $\times$ 4 (W) & order & $-$0.497 (0.008) & 1.000 & 1.000 & 0.292 & 0.509 \\
no signal (rqa) & 65 $\times$ 2, heterog.\ (B) & order & 0 (by arg.) & 1.000 & 0.000 & 0.429 & 0.493 \\
no signal (rqa) & 65 $\times$ 2, heterog.\ (B) & jitter 160 px & +0.001 (0.012) & 1.000 & 0.000 & 0.498 & 0.493 \\
no signal (rqa) & 65 $\times$ 2, heterog.\ (B) & jitter 40 px & +0.005 (0.011) & 1.000 & 0.000 & 0.451 & 0.493 \\
no signal (rqa) & 65, 16:49 (B) & order & 0 (by arg.) & 0.925 & 0.075 & 0.502 & 0.474 \\
\end{longtable}

%% file: supp_controls.tex
\begin{tabular}{lrrrrr}
\toprule
 & \multicolumn{2}{c}{Program contrast} & \multicolumn{3}{c}{Correctness, both programs} \\
\cmidrule(lr){2-3}\cmidrule(lr){4-6}
Descriptor & Real gaze & Synthetic gaze & All pairs & Perm.\ mean & Same-program pairs [95\% interval] \\
\midrule
Count & -- & -- & 0.495 & 0.515 & 0.517 [0.317, 0.646] ($p = 0.455$) \\
Dispersion & 0.957 & 1.000 & 0.716 & 0.710 & 0.490 [0.398, 0.581] ($p = 0.594$) \\
Classic & -- & -- & 0.501 & 0.505 & 0.523 [0.319, 0.625] ($p = 0.394$) \\
Dur.\ trend & -- & -- & 0.511 & 0.502 & 0.538 [0.373, 0.640] ($p = 0.262$) \\
RQA $0.5u$ & 0.580 & 0.604 & 0.571 & 0.517 & 0.527 [0.403, 0.645] ($p = 0.304$) \\
RQA $1u$ & 0.547 & 0.648 & 0.525 & 0.502 & 0.509 [0.384, 0.636] ($p = 0.462$) \\
Directed & 0.655 & 0.554 & 0.580 & 0.540 & 0.562 [0.410, 0.629] ($p = 0.130$) \\
FAG lattice & 0.658 & 0.859 & 0.664 & 0.566 & 0.577 [0.484, 0.685] ($p = 0.042$) \\
FAG AOI & 0.778 & 0.879 & 0.635 & 0.619 & 0.559 [0.441, 0.666] ($p = 0.087$) \\
FAG AOI pad. & 0.729 & 0.789 & 0.571 & 0.591 & 0.494 [0.345, 0.603] ($p = 0.575$) \\
\bottomrule
\end{tabular}

%% file: supp_expertise.tex
\begin{tabular}{lrrrrrr}
\toprule
Fold & \multicolumn{2}{c}{Classic} & \multicolumn{2}{c}{Directed} & \multicolumn{2}{c}{FAG AOI} \\
 & test & train & test & train & test & train \\
\midrule
1 & 0.686 & 0.532 & 0.400 & 0.593 & 0.229 & 0.588 \\
2 & 0.167 & 0.582 & 0.033 & 0.616 & 0.300 & 0.533 \\
3 & 0.143 & 0.595 & 0.381 & 0.589 & 0.095 & 0.598 \\
4 & 0.929 & 0.497 & 0.964 & 0.485 & 0.107 & 0.523 \\
5 & 0.429 & 0.551 & 0.257 & 0.594 & 0.200 & 0.595 \\
6 & 0.000 & 0.605 & 0.133 & 0.608 & 0.467 & 0.532 \\
7 & 0.300 & 0.562 & 0.500 & 0.576 & 0.100 & 0.549 \\
8 & 0.733 & 0.528 & 0.367 & 0.539 & 0.533 & 0.531 \\
9 & 0.214 & 0.533 & 0.214 & 0.538 & 0.286 & 0.542 \\
10 & 0.844 & 0.528 & 0.438 & 0.569 & 0.344 & 0.567 \\
\midrule
Mean & 0.444 & 0.551 & 0.369 & 0.571 & 0.266 & 0.556 \\
\bottomrule
\end{tabular}

%% file: supp_retention.tex
\begin{tabular}{lcccccc}
\toprule
Transformation & \texttt{em\_program} & \texttt{em\_halves} & \texttt{em\_expertise} & \texttt{em\_correct} & \texttt{em\_correctrect} & \texttt{em\_correctveh} \\
\midrule
none (observed) & 0.974 / 0.975 & 0.960 / 0.990 & 0.977 / 0.973 & 0.973 / 0.974 & 0.972 / 0.972 & 0.978 / 0.975 \\
order & 0.974 / 0.975 & 0.960 / 0.990 & 0.977 / 0.973 & 0.973 / 0.974 & 0.972 / 0.972 & 0.978 / 0.975 \\
duration coupling & 0.975 / 0.977 & 0.962 / 0.990 & 0.978 / 0.975 & 0.974 / 0.975 & 0.973 / 0.972 & 0.978 / 0.977 \\
duration marginal & 0.975 / 0.977 & 0.962 / 0.990 & 0.978 / 0.975 & 0.974 / 0.975 & 0.973 / 0.972 & 0.979 / 0.977 \\
rotation & 0.901 / 0.767 & 0.788 / 0.888 & 0.838 / 0.814 & 0.864 / 0.816 & 0.897 / 0.882 & 0.742 / 0.785 \\
jitter 0.25$u$ & 0.970 / 0.970 & 0.956 / 0.983 & 0.971 / 0.968 & 0.968 / 0.970 & 0.968 / 0.968 & 0.971 / 0.970 \\
jitter 0.5$u$ & 0.959 / 0.957 & 0.945 / 0.971 & 0.956 / 0.958 & 0.955 / 0.958 & 0.955 / 0.960 & 0.957 / 0.958 \\
jitter 1$u$ & 0.927 / 0.922 & 0.911 / 0.937 & 0.920 / 0.923 & 0.921 / 0.927 & 0.920 / 0.931 & 0.917 / 0.924 \\
jitter 2$u$ & 0.852 / 0.843 & 0.829 / 0.867 & 0.841 / 0.844 & 0.843 / 0.851 & 0.845 / 0.861 & 0.831 / 0.850 \\
\bottomrule
\end{tabular}

%% file: supp_matched.tex
\begin{tabular}{lllll}
\toprule
Problem & Contrast & AOI graph & AOI graph, support-matched & Support-only descriptor \\
\midrule
\texttt{em\_halves} (AUC 0.682; 0.773) & rotation & \textbf{+0.098 [+0.031, +0.166]} & +0.032 [$-$0.059, +0.124] (99\%) & +0.054 [$-$0.001, +0.110] \\
\texttt{em\_halves} (AUC 0.682; 0.773) & jitter 0.25$u$ & +0.036 [$-$0.029, +0.102] & +0.019 [$-$0.048, +0.086] (100\%) & \textbf{+0.048 [+0.018, +0.078]} \\
\texttt{em\_halves} (AUC 0.682; 0.773) & jitter 0.5$u$ & +0.023 [$-$0.025, +0.071] & +0.035 [$-$0.029, +0.099] (100\%) & \textbf{+0.112 [+0.065, +0.159]} \\
\texttt{em\_halves} (AUC 0.682; 0.773) & jitter 1$u$ & +0.045 [$-$0.032, +0.123] & +0.057 [$-$0.027, +0.141] (89\%) & \textbf{+0.119 [+0.059, +0.178]} \\
\texttt{em\_halves} (AUC 0.682; 0.773) & jitter 2$u$ & \textbf{+0.113 [+0.014, +0.212]} & \textbf{+0.121 [+0.035, +0.207]} (31\%) & \textbf{+0.123 [+0.054, +0.192]} \\
\bottomrule
\end{tabular}

%% file: supp_donor.tex
\begin{tabular}{lrrrrr}
\toprule
Problem & AUC obs. & AUC, LOO pool & AUC, training-fold pool (SD) & $\Cs$ LOO & $\Cs$ training-fold \\
\midrule
\texttt{gb\_tex\_vd1} & 0.969 & 0.935 & 0.935 (0.000) & +0.035 & +0.034 \\
\texttt{gb\_vd1\_blg} & 0.886 & 0.852 & 0.852 (0.001) & +0.033 & +0.034 \\
\texttt{gb\_vd1\_vd2} & 0.640 & 0.516 & 0.523 (0.009) & +0.124 & +0.117 \\
\texttt{gb\_tex\_s1s2} & 0.539 & 0.508 & 0.503 (0.020) & +0.031 & +0.036 \\
\texttt{gb\_vd1\_halves} & 0.564 & 0.556 & 0.553 (0.004) & +0.008 & +0.011 \\
\texttt{sb\_native} & 0.689 & 0.672 & 0.672 (0.002) & +0.017 & +0.017 \\
\texttt{sb\_compr} & 0.587 & 0.561 & 0.562 (0.005) & +0.026 & +0.024 \\
\texttt{em\_program} & 0.539 & 0.507 & 0.532 (0.010) & +0.032 & +0.007 \\
\texttt{em\_expertise} & 0.438 & 0.461 & 0.474 (0.038) & $-$0.023 & $-$0.037 \\
\texttt{em\_correct} & 0.501 & 0.447 & 0.454 (0.023) & +0.054 & +0.047 \\
\texttt{em\_correctrect} & 0.523 & 0.474 & 0.411 (0.050) & +0.048 & +0.112 \\
\texttt{em\_correctveh} & 0.500 & 0.475 & 0.454 (0.070) & +0.025 & +0.046 \\
\texttt{em\_halves} & 0.776 & 0.772 & 0.772 (0.005) & +0.004 & +0.004 \\
\bottomrule
\end{tabular}

%% file: supp_nulls.tex
\begin{tabular}{lllrrrrrr}
\toprule
Problem & Descriptor & Transformation & Observed AUC & $p$ & Transformed AUC & $p$ & Contrast & $p$ (lower tail) \\
\midrule
\texttt{em\_program} & Classic & order & 0.539 & 0.095 & 0.788 & 0.001 & $-0.249$ & 0.001 \\
\texttt{gb\_vd1\_halves} & Classic & order & 0.564 & 0.001 & 0.653 & 0.001 & $-0.090$ & 0.001 \\
\texttt{gb\_tex\_vd1} & Classic & order & 0.969 & 0.001 & 0.991 & 0.001 & $-0.022$ & 0.142 \\
\texttt{gb\_vd1\_blg} & Classic & order & 0.886 & 0.001 & 1.000 & 0.001 & $-0.114$ & 0.001 \\
\texttt{em\_halves} & RQA $0.5u$ & jitter $1u$ & 0.613 & 0.001 & 0.692 & 0.001 & $-0.080$ & 0.027 \\
\bottomrule
\end{tabular}

%% file: supp_big.tex
\begin{tabular}{lllrrrrr}
\toprule
Design & Descriptor & Contrast & Coverage & Excl.\ 0 at zero target & Width & SE ratio & Coverage at 80\% \\
\midrule
no signal & classic & order & 40/40 & 0/40 & 0.111 & 1.65 & 0.975 \\
no signal & directed & order & 38/40 & 2/40 & 0.139 & 1.39 & 0.950 \\
no signal & directed & jitter 160 px & 39/40 & -- & 0.148 & 1.26 & 0.900 \\
no signal & directed & jitter 40 px & 39/40 & -- & 0.150 & 1.39 & 0.975 \\
no signal & directed & rotation & 40/40 & -- & 0.144 & 1.42 & 0.900 \\
no signal & rqa & order & 40/40 & 0/40 & 0.133 & 1.63 & 0.950 \\
no signal & rqa & jitter 160 px & 40/40 & -- & 0.159 & 1.70 & 1.000 \\
no signal & rqa & jitter 40 px & 40/40 & -- & 0.150 & 1.62 & 0.925 \\
signal elsewhere & directed & order & 40/40 & 0/40 & 0.024 & 1.38 & 0.900 \\
signal elsewhere & directed & jitter 40 px & 40/40 & -- & 0.075 & 1.97 & 1.000 \\
signal elsewhere & directed & rotation & 40/40 & -- & 0.045 & 1.44 & 0.975 \\
signal elsewhere & rqa & order & 40/40 & 0/40 & 0.017 & 1.57 & 1.000 \\
signal elsewhere & rqa & jitter 40 px & 40/40 & -- & 0.070 & 1.81 & 0.975 \\
\bottomrule
\end{tabular}

%% file: supp_trees_long.tex
\begin{tabular}{lllll}
\toprule
Problem & Descriptor & Transformation & Trees, $R = 6$, $B = 60$ & Trees, $R = 20$, $B = 300$ \\
\midrule
\texttt{gb\_tex\_vd1} & Dur.\ trend & duration marginal & \textbf{+0.371 [+0.318, +0.424]} & \textbf{+0.378 [+0.327, +0.428]} \\
\texttt{gb\_vd1\_blg} & Classic & order & \textbf{$-$0.118 [$-$0.148, $-$0.088]} & \textbf{$-$0.118 [$-$0.146, $-$0.090]} \\
\texttt{gb\_vd1\_blg} & Dur.\ trend & duration marginal & \textbf{+0.226 [+0.161, +0.291]} & \textbf{+0.231 [+0.168, +0.293]} \\
\texttt{gb\_vd1\_vd2} & Classic & duration marginal & \textbf{+0.130 [+0.073, +0.187]} & \textbf{+0.127 [+0.074, +0.180]} \\
\texttt{gb\_vd1\_vd2} & Dispersion & jitter $2u$ & \textbf{+0.102 [+0.038, +0.166]} & \textbf{+0.100 [+0.036, +0.164]} \\
\texttt{gb\_vd1\_vd2} & Dur.\ trend & duration coupling & \textbf{+0.176 [+0.133, +0.219]} & \textbf{+0.172 [+0.128, +0.215]} \\
\texttt{gb\_vd1\_vd2} & Dur.\ trend & duration marginal & \textbf{+0.272 [+0.216, +0.329]} & \textbf{+0.274 [+0.226, +0.322]} \\
\texttt{gb\_vd1\_vd2} & Dur.\ trend & order & \textbf{+0.178 [+0.133, +0.224]} & \textbf{+0.175 [+0.132, +0.218]} \\
\texttt{gb\_vd1\_vd2} & RQA $0.5u$ & jitter $2u$ & \textbf{+0.106 [+0.041, +0.170]} & \textbf{+0.103 [+0.037, +0.168]} \\
\texttt{gb\_vd1\_vd2} & RQA $1u$ & jitter $2u$ & \textbf{+0.125 [+0.054, +0.195]} & \textbf{+0.123 [+0.067, +0.178]} \\
\texttt{gb\_vd1\_halves} & Directed & jitter $2u$ & \textbf{+0.192 [+0.140, +0.243]} & \textbf{+0.199 [+0.145, +0.252]} \\
\texttt{gb\_vd1\_halves} & Dispersion & jitter $2u$ & \textbf{+0.123 [+0.097, +0.150]} & \textbf{+0.124 [+0.096, +0.153]} \\
\texttt{gb\_vd1\_halves} & Dispersion & rotation & \textbf{+0.177 [+0.139, +0.215]} & \textbf{+0.167 [+0.133, +0.202]} \\
\texttt{gb\_vd1\_halves} & RQA $0.5u$ & jitter $2u$ & \textbf{+0.157 [+0.101, +0.214]} & \textbf{+0.165 [+0.108, +0.222]} \\
\texttt{gb\_vd1\_halves} & RQA $1u$ & jitter $2u$ & \textbf{+0.212 [+0.172, +0.253]} & \textbf{+0.219 [+0.168, +0.270]} \\
\texttt{sb\_native} & RQA $0.5u$ & jitter $2u$ & $-$0.002 [$-$0.084, +0.081] & +0.008 [$-$0.074, +0.091] \\
\texttt{sb\_native} & RQA $1u$ & jitter $2u$ & \textbf{+0.100 [+0.009, +0.190]} & \textbf{+0.111 [+0.024, +0.199]} \\
\texttt{sb\_native} & RQA $1u$ & jitter $1u$ & +0.071 [$-$0.003, +0.145] & \textbf{+0.082 [+0.012, +0.151]} \\
\texttt{em\_program} & Dispersion & rotation & \textbf{+0.166 [+0.088, +0.244]} & \textbf{+0.164 [+0.074, +0.255]} \\
\texttt{em\_halves} & Classic & order & \textbf{+0.128 [+0.033, +0.222]} & \textbf{+0.148 [+0.062, +0.233]} \\
\texttt{em\_halves} & RQA $0.5u$ & jitter $1u$ & $-$0.089 [$-$0.189, +0.011] & $-$0.087 [$-$0.192, +0.018] \\
\bottomrule
\end{tabular}

%% file: supp_descriptives.tex
\begin{tabular}{llrrrrrr}
\toprule
 & & \multicolumn{2}{c}{Trials} & \multicolumn{2}{c}{Participants} & \multicolumn{2}{c}{Fixations per trial} \\
\cmidrule(lr){3-4}\cmidrule(lr){5-6}\cmidrule(lr){7-8}
Problem & Positive class & positive & negative & positive & negative & median (IQR) & range \\
\midrule
\texttt{gb\_tex\_vd1} & \texttt{VD1} & 644 & 644 & 322 & 322 & 179 (132--225) & 37--300 \\
\texttt{gb\_vd1\_blg} & \texttt{BLG} & 643 & 644 & 322 & 322 & 121 (96--146) & 33--392 \\
\texttt{gb\_vd1\_vd2} & \texttt{VD2} & 644 & 644 & 322 & 322 & 141 (121--159) & 32--233 \\
\texttt{gb\_vd1\_halves} & \texttt{VD1\#2} & 644 & 644 & 322 & 322 & 66 (57--75) & 18--113 \\
\texttt{gb\_tex\_s1s2} & \texttt{TEX@S2} & 322 & 322 & 322 & 322 & 225 (206--243) & 97--300 \\
\texttt{sb\_native} & \texttt{nonnative} & 620 & 1426 & 29 & 66 & 122 (92--152) & 7--398 \\
\texttt{sb\_compr} & \texttt{low} & 805 & 558 & 79 & 66 & 121 (93--152) & 8--398 \\
\texttt{em\_program} & \texttt{vehicle} & 112 & 112 & 112 & 112 & 218 (145--291) & 25--868 \\
\texttt{em\_halves} & \texttt{second} & 224 & 224 & 112 & 112 & 109 (72--145) & 12--434 \\
\texttt{em\_expertise} & \texttt{low} & 64 & 51 & 37 & 28 & 181 (129--263) & 25--634 \\
\texttt{em\_correct} & \texttt{incorrect} & 147 & 124 & 120 & 108 & 212 (142--286) & 25--868 \\
\texttt{em\_correctrect} & \texttt{incorrect} & 41 & 92 & 41 & 92 & 193 (142--279) & 35--606 \\
\texttt{em\_correctveh} & \texttt{incorrect} & 106 & 32 & 106 & 32 & 230 (143--293) & 25--868 \\
\bottomrule
\end{tabular}